\documentclass{article}

\usepackage{PRIMEarxiv}

\usepackage[utf8]{inputenc} 
\usepackage[T1]{fontenc}    
\usepackage{hyperref}       
\usepackage{url}            
\usepackage{booktabs}       
\usepackage{amsfonts}       
\usepackage{nicefrac}       
\usepackage{microtype}      
\usepackage{lipsum}
\usepackage{fancyhdr}       
\usepackage{graphicx}       
\usepackage{subcaption}
\graphicspath{{media/}}     
\usepackage{comment}
\usepackage{enumitem}

\title{Investigating Temporal Convolutional
Neural Networks for Satellite Image Time
Series Classification: A survey
}

\author{
  James Brock \\
  Department of Engineering Mathematics \\
  University of Bristol \\
  Bristol\\
  \texttt{jb75426@gmail.com} \\
   \And
  Zahraa S. Abdallah \\
  Department of Engineering Mathematics \\
  University of Bristol \\
  Bristol\\
  \texttt{zahraa.abdallah@bristol.ac.uk} \\
}

\begin{document}
\maketitle

\abstract{Satellite Image Time Series (SITS) of the Earth's surface provide detailed land cover maps, with their quality in the spatial and temporal dimensions consistently improving. These image time series are integral for developing systems that aim to produce accurate, up-to-date land cover maps of the Earth's surface. Applications are wide-ranging, with notable examples including ecosystem mapping, vegetation process monitoring and anthropogenic land-use change tracking. Recently proposed methods for SITS classification have demonstrated respectable merit, but these methods tend to lack native mechanisms that exploit the temporal dimension of the data; commonly resulting in extensive data pre-processing contributing to prohibitively long training times. To overcome these shortcomings, Temporal CNNs have recently been employed for SITS classification tasks with encouraging results. This paper seeks to survey this method against a plethora of other contemporary methods for SITS classification to validate the existing findings in recent literature. Comprehensive experiments are carried out on two benchmark SITS datasets with the results demonstrating that Temporal CNNs display a superior performance to the comparative benchmark algorithms across both studied datasets, achieving accuracies of 95.0\% and 87.3\% respectively. Investigations into the Temporal CNN architecture also highlighted the non-trivial task of optimising the model for a new dataset.
}


\section{Introduction}\label{sec1}

\begin{figure}[ht]
\centering
\includegraphics[width=0.8\textwidth,height=\textheight,keepaspectratio]{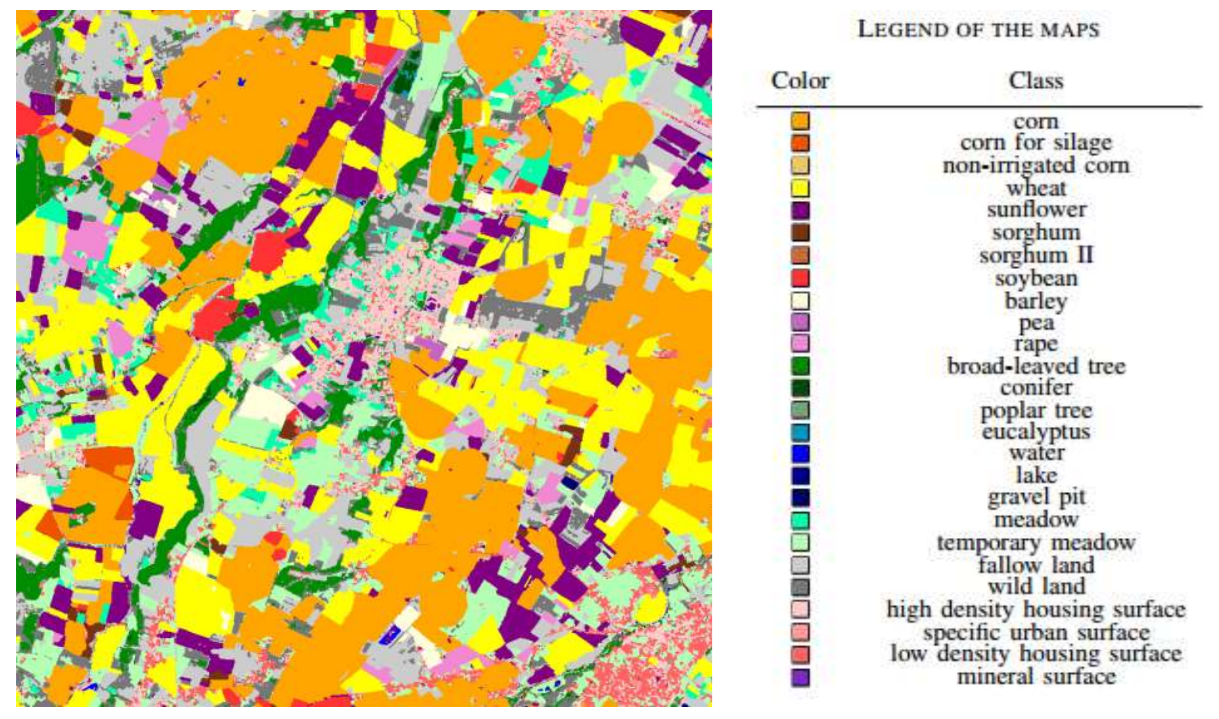}
\caption[SITS-TSI dataset overview]{An example Land cover map from the SITS-TSI dataset. Each pixel in the image is assigned to a target class. This is a snapshot of the dataset, with multiple images being stacked and geographically referenced to create the dataset. Image sourced from \cite{Tan2017}.}
\label{fig:Land cover map SITS-TSI}
\end{figure}

In recent decades there has been a dramatic increase in the availability of satellite-based earth imagery, resulting from \emph{Earth Observation (EO)} programs including the Sentinel and Landsat satellites \cite{schafer2018classifying}. Such satellites are able to capture images with spatial resolutions ranging from 10-100m with an increasing temporal density \cite{russwurm2019early,Pelletier2019}. These visual records of the Earth's ecosystems have existed since the 1970s but it has only been within the last decade that the data has been available for public use \cite{Pasquarella2016}. These datasets often contain channels of electromagnetic information not present in traditional 3-channel (\emph{e.g.} RGB) images, enabling a greater degree of data analysis. Whilst additional features are valuable for land-cover classification tasks, it consequently reduces the interpretability of the data and typically enlarges their volume \cite{SainteFareGarnot2020}. An example land-cover map for a SITS classification task can be observed in Figure \ref{fig:Land cover map SITS-TSI}. 

The advent of modern computing power and public data availability has opened the avenue for the wider scientific community to utilise land-cover maps within studies of limited funding \cite{schafer2018classifying}. Most notably, this data has proven an essential asset for wide-scale ecological mapping and monitoring; a key tool for shaping future climate change policies. Indeed, the extensive multi-spectral bands offered by \emph{EO} imagery enable accurate and comprehensive mapping of the planet for the evaluation of forest fire spread, land-use change and vegetation cover \cite{COP26SatelliteBriefing}. Increasingly more traditional disciplines such as ecology and biology are coming to rely on land-cover maps to evaluate ecosystems and help inform public policies \cite{Running652}; current applications of SITS classification have included agricultural crop monitoring \cite{Zhong2019,russwurm2019early}, mapping dead forest cover \cite{Sylvain2019}, monitoring tropical forest succession patterns \cite{caughlin2021monitoring}, mapping plant communities \cite{Rapinel2019}, and wider ecosystem monitoring as a whole \cite{Guo2014}. With many sites being difficult to access and survey on the ground, satellite imagery offers a cheaper and more reliable alternative to previous mediums, with studies utilising, SITS data often having reduced costs, fastest processing times and more accurate results.

This paper primarily seeks to demonstrate the viability of Temporal CNNs for SITS classification by validating the results of prior studies - focusing on the application of convolutions on the temporal channel and the exploitation of spectral features.  By performing experiments on two datasets with seasonal land-cover labels, the performance of the Temporal CNN and other surveyed methods across disparate feature spaces and temporal length will be thoroughly explored. The contributions of this paper can be summarised as follows:
\begin{itemize}
    \item Providing a comprehensive analysis of the recently proposed Temporal CNN architecture \cite{Pelletier2019}, comparing it against five other state-of-the-art architectures for STIS classification - including how well each model generalises across the studied datasets
  \item Evaluating how sensitive the Temporal CNN's architectural parameters are to the features of each dataset - focusing on the effects of changes within the temporal and spectral channels
  \item Identifying relationships between dataset attributes and their influence on optimal hyperparameter configuration which can be used to more rapidly discern the optimal architecture for datasets with specific characteristics.
\end{itemize}

The experiments conducted in this paper were designed to incorporate many of the recommendations found in the literature relevant to Temporal CNNs for SITS classification. However, this paper still heavily investigates the optimal parameter settings for aspects of the model to validate prior results. This investigation includes an analysis of the optimal width and depth of the model, the degree of temporal dilation required, and the extent to which regularisation is required. By conducting experiments on two datasets, a more generalised investigation was able to provide insights into the degree to which data composition affected the selection of an optimal model. An extensive study is then conducted using these optimal parameters for the Temporal CNN which compares the model to other contemporary algorithms for SITS classification. In this comparison, hyperparameter searches were conducted for each architecture on both datasets in order to obtain a fair comparison between the optimal models.

The remainder of this paper is structured as follows: Section \ref{related work} provides background on past and current methods for time series and SITS classification. Section \ref{background} then discusses the fundamental components of the Temporal CNN. Next, Section \ref{surveyed methods} describes each of the surveyed methods explored within this paper, detailing their unique architectural components and composition. Section \ref{DS} outlines the datasets used in this study, details of pre-processing steps applied, and the performance metrics are chosen. Section \ref{exp} comprises the bulk of the paper, containing details of the experimental setup, experiments investigating the optimal architecture and degree of regularisation for the Temporal CNN on both datasets and the results comparing the Temporal CNN to the optimised surveyed models. Finally, Section \ref{conclusion} concludes the paper with a critical discussion of the results, their subsequent conclusions and future work.

\subsection{A note on terminology}
For the context of this paper, it is useful to note the terminology for the various components related to the SITS classification domain. SITS is an abbreviation for \textit{Satellite Image Time Series} which are series of images collected over a prolonged time period via satellites using optical (e.g. Sentinel-2) and radar (Sentinel-1) sensors which produce Earth observation data \cite{6144005}. These images are multi-spectral in nature, meaning that they are composed of a series of sensors operating in various bands of the electromagnetic spectrum. By extension, this also means that the data is multi-variate if more than one sensor value constitutes an observation. This data is therefore comprised of temporal components integrated with spectral and spatial dimensions, inherently making it 4D in nature (spatial components (x, y) + temporal + spectral). SITS datasets focus on documenting changes in the land cover of an area by monitoring and mapping the differences between the various temporal snapshots. Given this, SITS classification is a form of multivariate time series classification; multi-variate in the sense of temporal snapshots comprised of spectral and spatial information. This data forms highly detailed remotely sensed land-cover maps that can be used for classification and change detection tasks through the employment of various time series analysis techniques. For processing this data, deep neural networks have gained in popularity within contemporary literature and are subsequently referred to as DNNs throughout the rest of this paper.

\section{Related Work}
\label{related work}
The availability and quality of SITS data have generated a renewed interest in discovering more efficient classification algorithms through which to process the data. Previous studies have demonstrated good performances for SITS classification when using traditional approaches \cite{Gomez2016}, but these approaches commonly lack intrinsic methods of processing the temporal dimension of the data. Conversely, various deep learning methods have recently been studied that contain mechanisms for processing the temporal dimension of the data whilst discovering meaningful features within the data \cite{Pelletier2019,IsmailFawaz2019}. Given the vast areas for which high-quality SITS data now exists, it is imperative that deep learning is thoroughly explored for the task so researchers can learn how best to apply it.

\subsection{Supervised learning for SITS classification}
Traditional approaches have commonly employed models that require features to be manually crafted, such as the Normalised Difference Vegetation Index (NDVI) \cite{Reddy2018}. Contemporary papers that introduce a new method for time series land-cover classification commonly compare performance with supervised classifiers such as Support Vector Machines (SVMs) \cite{Pelletier2016}, Random Forests (RFs) \cite{Lopes2020}, and recurrent neural network (RNNs) variants \cite{Russwurm2020}. RNNs and ensemble methods such as RFs have consistently proven to be strong baselines for remote-sensing classification tasks \cite{Zhong2019}, resulting in their continued adoption. Of these methods, Random Forests and SVMs have no mechanisms for processing the temporal dimension of the data and so valuable information provided by seasonal changes is not utilised \cite{Pelletier2019}.  Previous studies incorporating these methods include mapping dead tree cover \cite{Sylvain2019}, mapping floodplain grassland communities \cite{Rapinel2019}, monitoring ecosystem states and dynamics \cite{Guo2014} and agricultural crop analysis \cite{pena2015assessing}.

Comprehensive reviews of traditional classifiers for Time Series Classification (TSC) report that machine learning and ensemble algorithms are more accurate and efficient than conventional classifiers such as SVMs when faced with high-dimensional, complex datasets \cite{Gomez2016, moumni2021new}. The reviews by Fawaz et al \cite{IsmailFawaz2019} and Moskolai et al \cite{moskolai2021application} avidly emphasises the potential that deep learning models have in the TSC domain; with CNN models routinely demonstrating strong performances. Whilst more traditional supervised learning methods can often deliver respectable performances, their main drawback is the requirement for manual complex feature engineering which has a limited capacity for automation and is characteristically rigid \cite{Zhong2019}. Caveats in prior reviews comparing neural network methods versus RFs and SVMs displayed that RFs and SVMs are more competitive in tasks where fewer data samples are available \cite{liu2018comparing}. For the purposes of this study though the number of available data samples is not a limitation that would prevent a fair comparison between these sets of methods.

Each of these non-parametric methods generally does not require large amounts of training data to achieve a good performance; although SVMs struggle in small feature spaces \cite{khatami2016meta}. Each method is computationally intense, with NNs and RFs acting as black boxes, making the interpretation of results difficult despite RFs having the capacity to determine variable importance \cite{Pelletier2016}. Lastly, each method - with the exception of the Random Forest - requires extensively tuned input parameters for optimal performance, which may exacerbate the already long training times \cite{Gomez2016}; especially for models that cannot be trained in parallel. Setting the parameters of RFs has previously been found to have little influence on their classification accuracy for SITS tasks \cite{Pelletier2016}, and hence can be employed effectively with little hyperparameter tuning. If over-fitting is overcome, then NNs generally perform the best since feature representations are learnt automatically, reducing the required pre-processing needed for model development \cite{Russwurm2020}, although the discovery of a well-performing model is not trivial. When fairly assessing the performance of these various methods with regard to their training times, the ability for each model to be trained in parallel needs to be considered a factor. For example, the decision trees in an RF can be trained in parallel across multiple worker threads which can dramatically shorten training times depending on the computational resources available \cite{senagi2017using}.  

\subsection{Deep learning approaches for SITS}


Recently, a wide-ranging variety of deep learning models have been applied to SITS tasks and time series classification (TSC) as a whole. Reviews such as those conducted by Fawaz et al. \cite{IsmailFawaz2019}, Hatami et al. \cite{hatami2018classification} and Campos-Taberner et al. \cite{campos2020understanding} methodically compare the most notable methodologies for both SITS and TSC applications; being tested on benchmarks such as the UCR archive \cite{dau2019ucr} as well as manually curated SITS datasets. The most common forms of DNNs for SITS classification are summarised below in Table \ref{tab:SITS-Classification-Deep-Learning-Methods}. As observed in papers like \cite{Ienco2019}, the most common architectures are generally a choice between CNNs \cite{Sylvain2019,Cui2016}, RNN varieties such as Bi-LSTMs \cite{campos2020understanding} and GRUs \cite{russwurm2018multi}, and most recently more novel approaches such as Transformer models that have been adapted from the language modelling domain \cite{vaswani2017attention}. Recently, there has been a plethora of hybrid architectures proposed \cite{russwurm2018convolutional, garnot2019time, xu2021novel} which are outside of the scope of discussion of this paper, by the review by Moskola{\"\i} et al. \cite{moskolai2021application} provides an excellent overview of some of these methods. Two recent papers that introduced Transformer models for SITS classification \cite{SainteFareGarnot2020, Garnot2020} were shown to outperform the Temporal CNN model of Pelletier et al. \cite{Pelletier2019}, but they perform classification on a per-parcel basis instead of a per-pixel basis, making them incompatible for a comparison in this study. 
 
\begin{table}[ht!]
\centering
\begin{tabular}{ |p{0.15\linewidth}|p{0.26\linewidth}|p{0.52\linewidth}| }
 \hline
 \textbf{Architectural type} & \textbf{Notable models} & \textbf{Key findings} \\
 \hline
 CNNs & Temporal CNN \cite{Pelletier2019}, InceptionTime \cite{IsmailFawaz2020}, Time-CNN \cite{Zhao2017}, MCDCNN \cite{zheng2016exploiting, zheng2014time}, 2D CNN \cite{simon2022convolutional} & \begin{itemize}[leftmargin=.2in]
  \item Ability to extract spatial features within 2D spaces such as images \cite{krizhevsky2012imagenet} which naturally translates well into the SITS domain
  \item 1D, 2D and 3D ConvNets have been developed that are capable of processing combinations of the temporal, spatial and spectral dimensions \cite{Zhao2017}
  \item Operations of CNNs can be parallelized, giving it an edge over classic RNNs, despite the high computational cost convolutional operations incur
  \item Convolutional filter size can be adjusted depending on the temporal length and sampling of a dataset for increased flexibility
  \item Important to limit the number of trainable parameters in a CNN architecture to reduce training times and the amount of data required 
  \end{itemize}\\
 \hline
 RNNs & RNN-LSTM \cite{Russwurm2020}, GRU \cite{garnot2019time, chung2014empirical}, StarRNN \cite{russwurm2019breizhcrops}, DeepCropMapping \cite{xu2020deepcropmapping} & \begin{itemize}[leftmargin=.2in]
  \item Inherent nature for processing sequential data allows RNNs to fully exploit the temporal dimension in SITS datasets \cite{Zhong2019}
  \item Performance of RNNs degrade on longer sequences due to the vanishing gradient problem. Bi-LSTMs and GRUs limited this problem by introducing specialised components that could represent the temporal dependency at various time spans with gated recurrent connections \cite{Reddy2018}
  \item Baseline LSTMs have been extended to make use of 2D convolutional layers to act as spatial feature extractors that provide inputs to the LSTM \cite{mou2018learning}; demonstrating the value of employing both the temporal and spatial channels \cite{Zhong2019}.
  \end{itemize} \\
 \hline
 Transformers & PSAE+TAE \cite{SainteFareGarnot2020}, PSE + L-TAE \cite{Garnot2020}, Transformer \cite{Russwurm2020}, Informer \cite{yan2022land}, GL-TAE \cite{zhang2023attention} &  \begin{itemize}[leftmargin=.2in]
  \item Utilise mechanisms such as self-attention combined with positional encoding \cite{Russwurm2020} or pixel set-encoders to encode the spatial context of the data in collaboration with a temporal attention encoder to encode the temporal relations between observations \cite{SainteFareGarnot2020, Garnot2020}.
  \item Shown to offer state-of-the-art performances across a range of spatial resolutions
  \item Particularly suited to classifying data that has undergone little to no-preprocessing \cite{Russwurm2020}, a positive point give how expensive and tedious data pre-processing can be for SITS datasets
  \end{itemize} \\
 \hline
\end{tabular}
\caption[Comparison of deep learning approaches for SITS classification]{\label{tab:SITS-Classification-Deep-Learning-Methods}Overview of previously employed deep learning techniques for SITS classification and their key findings. }
\end{table}

\subsection{Temporal CNNs}
Building from the earlier discussion on CNNs for TSC tasks, CNNs of various dimensional capacities (1D, 2D, 3D) have been applied to time series data. 1D (spectral convolutions) and 2D (spatial convolutions) CNNs have demonstrated their effectiveness for multi-source, multi-temporal datasets, but did not take advantage of the temporal dimension \cite{kussul2017}. The convolutions are either applied in the spectral or spatial domain, excluding the temporal domain. As a result, the order of the images has no influence on classification, removing an essential component for applications that use features such as vegetation growth patterns \cite{Pelletier2019} to aid the classification of crops for example. In light of this limitation, a series of alternative methodologies have been proposed, including ensemble methods \cite{Abdallah2020}, methods derived from statistical analysis \cite{schafer2018classifying}, pixel-set encoders with temporal attention encoders \cite{Garnot2020, SainteFareGarnot2020}, self-attention with positional encoding \cite{Russwurm2020}, and most notably for this paper, CNNs that incorporate the temporal channel \cite{Pelletier2019,Zhao2017}.

Temporal Convolutional Neural Networks (Temporal CNNs) were first introduced for action segmentation \cite{lea2016temporal} and object detection within videos \cite{lea2017temporal}, achieving competitive or superior performance to comparable methods whilst boasting a significant reduction in training time. Since then, they have been expanded into the domain of sequence modelling, most notably for SITS classification using crop data \cite{Pelletier2019, ravcivc2020application}. The deep learning approach applies convolutions in the temporal dimension in order to automatically learn the temporal and spectral features of a dataset.

Given the structural similarities of video data to SITS and the promising results Temporal CNNs have displayed for SITS classification \cite{Zhong2019} and remote sensing \cite{Wang2017}, the choice to investigate this methodology is markedly justified. Whilst more traditional and generic 1D and 2D-CNNs have the ability to apply convolutions to the temporal dimensions, Temporal CNN architectures are designed specifically to fully exploit the temporal structure of SITS. 3D-CNNs have also shown promise for handling the spatial and temporal dimensions in video classification tasks, which could feasibly be adapted to SITS classification \cite{Ji2018}. The aforementioned works highlight the potential of applying Temporal CNNs for SITS classification, with Temporal CNNs possessing higher accuracies and shorter training times when compared to more traditional approaches such as RFs and RNNs \cite{Pelletier2019}. 

Temporal CNNs are a significant breakthrough for SITS tasks, with their prominence for handling temporal data well documented \cite{ravcivc2020application}. They are able to match and surpass more traditional methods such as RNNs and RFs by using an end-to-end deep learning architecture with significantly reduced training times \cite{bai2018empirical}. Results from papers applying Temporal CNNs to SITS classification tasks such as \cite{Pelletier2019} achieved overall accuracies of between 93-94\% \cite{Pelletier2019}, outclassing RNNs and Random Forests by 1-3\%. Another paper that also applied Temporal CNNs to SITS data in the form of a 2D CNN \cite{debella2021mapping} achieved overall accuracies between 88\% and 94\%. The 2D CNN was implemented in the case where the data are conceptualized as 2D data, with the spectral and temporal dimensions taking one dimension each. A 2D kernel (spectral and temporal dimensions) is then used to convolve the data to extract various features \cite{debella2021mapping}. A study by \cite{ravcivc2020application} compared two Temporal CNN models: TempCNN \cite{Pelletier2019} and TCN \cite{bai2018empirical} on a SITS classification task, a task the TCN at the time not been tested on. Both methods were comparable, with the TCN architecture offering an alternative in circumstances with limited data, computing power or time \cite{ravcivc2020application}. For this study, the Temporal CNN is a preferred choice for experimentation and analysis.

\section{Temporal CNN Fundamentals}
\label{background}
This section aims at presenting Temporal CNN models and their characteristic components. The theory for Temporal CNNs is introduced, followed by an overview of the Temporal CNN architecture that is experimented on in section \ref{exp} is introduced. Appendix \ref{secA1} comprises an overview of general deep learning concepts that may be used as a prerequisite to this section for readers unfamiliar with deep learning.

\subsection{Convolutional layers}
Convolutional layers are the characteristic component of CNNs. Convolutional networks combine three architectural mechanics to ensure a level of invariance to shifts and distortions: local receptive fields, shared weights and less commonly spatial or temporal sampling \cite{lecun1995convolutional}. A convolutional neural network achieves this through the use of convolutional layers. Convolutional layers take the output of the previous layer and extract features by correlating the received input with a set of convolutional filters (kernels) that apply convolutional operations which are followed by an elementwise nonlinear activation function $\phi$ \cite{Russwurm2020}. The size of the convolutional filter used determines the size of the receptive field, with smaller filter sizes resulting in a more localised neighbourhood of units being used as an input into the next layer \cite{lecun1995convolutional}. The size of the receptive field increases through the number of layers, subsequently combining the features extracted within local receptive fields at the higher levels which creates feature maps that can be correlated with specific output classes \cite{Russwurm2020}. It naturally follows that any locally extracted features that are useful in one part of the input will likely be useful across the rest of the input; this is the realisation of weight sharing. Weights are shared by forcing a set of units - whose receptive fields are located within different parts of the input - to share identical weight vectors by sliding the collection of weights across the entire input \cite{lecun1995convolutional, Pelletier2019}. This ability to share weights dramatically reduces the number of weights in the layer. Consequently, the number of trainable parameters in the network depends only on the filter size used and the number of units in the network; not the size of the input \cite{Pelletier2019}. The output of a convolutional layer however is dependent on the size of the input and the values chosen for the stride and padding. The stride refers to the distance between the receptive field centres of neighbouring neurons in a kernel map \cite{krizhevsky2012imagenet}. Padding is used to help the feature extractor visit information at the edges of the input by padding the input with neutral values such as 0, ensuring the original information is unaffected \cite{hashemi2019enlarging}.

Across the input, different types of units will discover different feature maps which will compute a variety of features. The core purpose of a convolutional layer is to automatically extract features from the input through the generation of feature maps. Once a feature has been detected, its exact location becomes less important, rather than its position in relation to other features is preserved \cite{lecun1995convolutional}. Accordingly, it is common practice to place layers after the convolutional layer that perform local averaging, normalisation, and provide non-linear activations which contribute to reducing the resolution of the feature maps but reduces the sensitivity of the model to noise \cite{Cui2016, russwurm2019breizhcrops}. 

\subsection{Temporal convolutions}
The temporal CNN as introduced by \cite{Pelletier2019} is fundamentally a variety of CNN that applies 1D convolutions along the temporal axis of the dataset to exploit the sequential nature of the observations. In the case of a time series, a 1D CNN applies a moving window to the series, which incorporates the temporal information in the classification process by extracting temporal features such as crop growth patterns \cite{simoes2021satellite}. It is important to clarify here that the spectral information contained within each observation provides crucial assistance in classification, e.g. grey patches representing urban class labels.

\begin{figure*}[ht!]
\centering
\includegraphics[width=1\textwidth,height=\textheight,keepaspectratio]{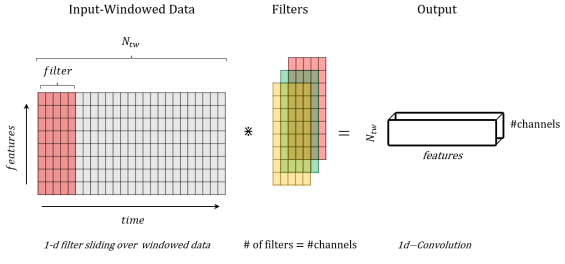}
\caption[1D Convolution diagram]{In the case of temporal data, convolution filters are applied along the length of the time series, applying filters along the available channels at each time step. In the context of SITS classification, spectral information and possibly spatial information (if available) are used to build feature maps at each time step to aid classification. Image sourced from \cite{chao2022fusing}.}
\label{fig:1D_Convolution}
\end{figure*}

Time series naturally have a strong 1D structure, which in the case of this SITS application can be exploited since pixels that are spectrally similar or temporally nearby are highly correlated. Using the inherent nature of convolutional networks to extract local features through small receptive fields ensures that such local correlations can be discovered and subsequently used to identify characteristics for classification, e.g. crop growth patterns \cite{lecun1995convolutional, Zhong2019}. In temporal applications, the size of the convolutional filter will affect how many timestamps the receptive field can process, with later convolutional layers having a greater temporal reach \cite{simoes2021satellite}. Generally, the longer the time series, the larger the filter size becomes since the goal is to discover a seasonal variation within the data \cite{hatami2018classification, IsmailFawaz2020}; this idea has recently been applied successfully to remote sensing applications \cite{Zhong2019, Pelletier2019}.  Figure \ref{fig:1D_Convolution} provides a useful graphical overview of how 1D convolutions can be applied to a time series. 

\section{State-of-the-art methods in SITS}
\label{surveyed methods}
This section reviews a number of benchmark algorithms to offer a comprehensive overview of the methodologies available for SITS classification in contrast to the Temporal CNN. These algorithms are now briefly introduced, along with the parameters that are modified for each model as part of the contribution of this paper. To adapt each of the architectures to the dataset in use, it was common to provide them with the temporal length, the number of features per timestamp and the number of output classes being predicted.

To reduce the required search for hyperparameter optimisation due to computing limitations, a search for the optimal learning rate $\nu$ and weight decay $\lambda$ was not performed here. Instead, for each model on both datasets values were taken from papers that have made prior recommendations as a result of their own extensive optimisations on multivariate time series classification tasks. Due to higher dropout values decreasing model performance and no longer reducing overfitting, the dropout value search space for all models is $d ~ ([0, 0.5])$. Naturally, the tuneable hyperparameters for each model varied and are listed below in each model's subsection as part of the contribution of this paper. The optimised values for all of these hyperparameter values are found in Section \ref{optimal_settings}; apart from those for the Temporal CNN which are discussed in Section \ref{TCNN_Experiments}.

\subsection{Temporal CNN network}
\begin{figure*}[ht!]
\centering
\includegraphics[width=\textwidth,height=\textheight,keepaspectratio]{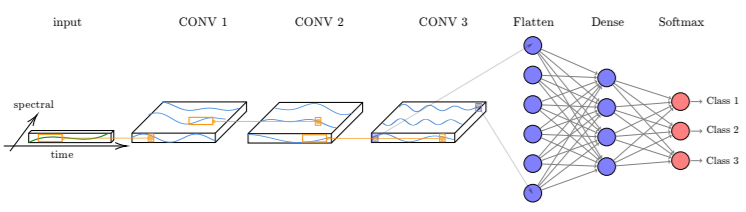}
\caption[Experimental Temporal CNN architecture]{The input to the architecture is a multivariate time series; SITS in this case. Three consecutive convolutional layers are applied that each applying convolutions along the temporal dimension, processing the spectral information at each timestamp. This is then followed by one dense layer and a final Softmax layer that produces the output class predictions. Image sourced from \cite{Pelletier2019}.}
\label{fig:Temporal_CNN_Architecture}
\end{figure*}

The architecture being investigated is that proposed by Pelletier et al. \cite{Pelletier2019}, which has seen recent usage in \cite{russwurm2019breizhcrops, Russwurm2020}. Figure~\ref{fig:Temporal_CNN_Architecture} displays the Temporal CNN architecture that this paper experiments on. The architecture is composed of three convolutional layers with 64 units, one dense layer of 256 units and one Softmax layer of N neurons for a classification task of N output classes. Their experiments found a maximum accuracy with a filter size of 9, but 5 is used to strike a balance between model complexity and performance, and as such is used here by default in the experiments \cite{russwurm2019breizhcrops}. The outputs of all layers except the Softmax layer employ batch normalisation and ReLU operations. They found the optimal dropout rate to be 0.182 \cite{russwurm2019breizhcrops}.

This paper contributes in Section \ref{TCNN_Experiments} extensive hyperparameter tuning on the architectural components mentioned above across the two studied datasets. Specifically, the number of convolutional layers and the number of neurons in them, the number of neurons in the dense layer, the filter size, batch size and dropout. By observing any differences in the optimal hyperparameter settings between the two datasets, sensitivities amongst specific dataset features can be identified. For the experiments, the validation set size was selected to be 10\% rather than 5\% to help improve the clarification of the cross-validation results. Weight decay is kept at the default value of $10^{-6}$ as it was demonstrated to have a negligible effect on model performance \cite{Pelletier2019}. The parameters of the network are trained using the Adam optimiser \cite{kingma2014adam} with default settings ($\nu = 0.001, \lambda = 0.0$), using a batch size of 128 and drop out of 0.182 \cite{russwurm2019breizhcrops}. Once the optimal hyperparameters are found for each model, they are retrained on enough epochs to allow the early stopping mechanism to finish training; producing models with minimal overfitting and optimal performance. 

\subsection{Multi-Channel Deep Convolutional Neural Network}
The methodology proposed by Zheng et al. \cite{zheng2014time,zheng2016exploiting} introduced a multi-channel deep convolutional neural network (MCDCNN) for time series classification. The aim of the architecture is to exploit presumed independence between the different features of the multi-modal time series data by applying the convolutions independently (\emph{i.e.} parallel) on each dimension of the input. Given a multivariate time series, the architecture first learns features from the individual univariate time series in each channel, which are then concatenated together to create an overall feature representation that is fed to the final layer. Classification is performed through a Multilayer Perceptron (MLP) to calculate class probabilities. Predictions are generated using the categorical cross-entropy loss function paired with a Softmax output layer and an Adam optimiser.

The process can be described as follows: each dimension for an input multivariate TS goes through two convolutional stages using 8 filters of length 5 with a ReLU activation function. Each convolution is followed by a max pooling operation using a kernel of size 2. The outputs of the second convolution for each dimension are then concatenated together which is then fed to a fully-connected layer with 732 neurons with ReLU as the activation function. Two more dense layers follow with 256 and 128 neurons respectively, the first layer is followed by a dropout layer with a dropout of 0.3, and the second layer's output is passed to a batch normalisation layer. A Softmax classifier is then used to make predictions with the layer containing a number of neurons equal to the number of classes \cite{zheng2016exploiting}. The model as given by \cite{IsmailFawaz2019} is compiled with an SGC optimiser and used default values for $\nu$ and $\lambda$ as 0.01 and 0.0005 respectively. This paper only adjusts the dropout rate to find the optimal model.

\subsection{Time-CNN}
The Time-CNN as introduced by Zhao et al. \cite{Zhao2017} seeks to build upon the work of the MCDCNN \cite{zheng2014time,zheng2016exploiting} by addressing the limitations they saw within the model. They argue that the model cannot learn the relationships between the various univariate time series that compose the multi-modal data since the model processes the data as multiple univariate time series and not a single multivariate time series. This allows the model to instead alternatively apply convolution and pooling operations to the multivariate time series to generate deep features of the raw data, jointly training the multivariate time series to extract features concurrently instead of independently. These features are then connected to a multilayer perceptron (MLP) to perform classification. The authors hoped that instead of learning the features from individual univariate time series that could not be correlated together effectively, the new joint training on the multivariate time series would produce feature maps that would improve and provide more accurate class predictions.

Further experimentation conducted by them investigated the optimal convolutional filter size, the type of pooling layer and the number of convolutional filters used. This experimentation resulted in the replacement of the categorical cross-entropy loss and Softmax output layer with an MSE loss and sigmoid output layer, average pooling in place of max pooling, and omitting the use of a pooling layer after the last convolutional layer. Specifically, the network as described in \cite{Zhao2017, martino_2020} is composed of two consecutive convolutional layers that use a filter size of 5, using 6 and 12 filters respectively, which are separated by a local average pooling operation with a filter size of 3. The sigmoid activation function is adopted as the activation function of the convolutional layers. After the second convolutional layer, the outputs are flattened and fed to a series of three dense layers with the first layer being followed by a dropout layer and the second layer followed by a batch normalisation layer. The dense layers have 128, 64, and 32 neurons respectively and the dropout rate used is 0.3. The output layer consists of a fully-connected layer with neurons equal to the number of output classes \cite{IsmailFawaz2019}. The model is compiled with an Adam optimiser with default values for $\nu$ and $\lambda$ set at 0.001 and 0.0 respectively. Like the MCDCNN, the dropout rate is again the only hyperparameter investigated when finding the optimal model.

\subsection{Recurrent Neural Network (RNN)}
Recurrent Neural Networks are regularly used as a comparative baseline within deep learning research for time series classification \cite{SainteFareGarnot2020,bai2018empirical}. These networks share the learned features across different positions within a sequence; intuitively modelling the temporal dependencies of the data. However, as the error is back-propagated at each time step, the computational cost can become significant and cause potential learning issues such as the vanishing gradient problem, in which inputs earlier in the sequence are forgotten. More recent RNN architectures use LSTM and GRU units that help to capture the long-distance connections and solve the vanishing gradient problem \cite{Pelletier2019}. The specific RNN architecture used in this comparison is the one proposed by Pelletier et al. \cite{Pelletier2019} which uses three stacked bidirectional GRUs (128 neurons), one dense layer (256 neurons) and a Softmax output layer using an L2 rate of $10^{-6}$ with an Adam optimiser using $nu = 0.001$ and $\lambda = 0.0$ and dropout of 0.571 \cite{russwurm2019breizhcrops}. This paper contributes an investigation into the optimal values for the following hyperparameters: the number of cascaded layers over a search space of $L \in {1,2,3,4}$, hidden vector dimensions of $H \in {2^6, 2^7, 2^8}$, fully connected units of $F \in {2^6, 2^7, 2^8, 2^9}$, and the dropout value. Multiple models were developed for both datasets with only the best results on the test data being reported. Early stopping was used to reduce overfitting with the validation loss being the monitored metric. 

\subsection{InceptionTime}
InceptionTime \cite{IsmailFawaz2020} is an ensemble of five deep CNN models specialised for image recognition tasks, being most heavily inspired by the Inception-v4 CNN network \cite{szegedy2016inceptionv4}. The model takes inspiration from the revolutionary AlexNet model; adapting it for the TSC domain.  It is reported in the introductory paper that the model outperforms the previous state-of-the-art models such as HIVE-COTE \cite{lines2016hive} and ResNet \cite{Wang2017} and does so whilst being much more scalable, having the ability to learn from an unprecedented quantity of training data within a short time frame. This model has demonstrated strong performance for SITS classification tasks having previously been applied to the SITS-TSI dataset \cite{Tan2017}. The core of the InceptionTime model is the Inception module which works by applying several filters of various resolutions to the input multi-modal time series data.\\

The Inception module of the network is composed of the following layers:
\begin{itemize}
    \item A bottleneck layer (32 neurons) to reduce the dimensionality (\emph{i.e.} depth) of the inputs, consequently cutting the computational cost by reducing the number of parameters, speeding up training and improving generalisation.
    \item  The output of the bottleneck is fed to three sequential 1D convolutional layers of kernel sizes 10, 20 and 40 that each employ 32 filters by default.
    \item The input of the Inception module is passed through a max pooling layer of size 3 and then through a bottleneck layer (32 neurons).
    \item The last layer is a depth concatenation layer where the output of the bottleneck layer in the step prior is concatenated along the depth dimension.
\end{itemize}

Given that the InceptionTime network has been shown to outperform the ResNet network \cite{IsmailFawaz2020}, which itself had outperformed the MCDCNN \cite{zheng2016exploiting} and Time-CNN \cite{Zhao2017} models in multiple studies on TSC tasks \cite{fawaz2019deep, IsmailFawaz2019}, it makes a strong case that InceptionTime should itself, in turn, surpass the performance of these two models, with the later experimental results looking to reflect this. The values for $\nu$ and $\lambda$ are set to default for the Adam optimiser with 0.001 and 0.0 respectively. The learning rate was reduced by a factor of 0.5 each time the model’s training loss has not improved for 50 consecutive epochs (with a minimum value equal to 0.0001). In this paper, the stacking of $L \in {3,4,5,6}$ Inception modules with $H \in {2^6, 2^7, 2^8}$ hidden units are experimented with, alongside the dropout value.

\subsection{Transformer network}
Given the recent success of transformer-based models in the SITS domain \cite{Russwurm2020}, it is appropriate to introduce them here as a comparative model. The transformer model used is a PyTorch implementation provided by Ru{\ss}wurm et al. \cite{russwurm2019breizhcrops}, which implements the transformer model presented in Ru{\ss}wurm and K{\"{o}}rner \cite{Russwurm2020}. The experimental results for the Transformer model in \cite{russwurm2019breizhcrops} are similar to those of \cite{Russwurm2020}, with the Transformer offering competitive performances to the Temporal CNN, InceptionTime and LTSM models - setting a strong comparative baseline for what can be expected in this study.

The model developed by Ru{\ss}wurm and K{\"{o}}rner \cite{Russwurm2020} is a SITS-adapted version of the original Transformer model introduced in \cite{vaswani2017attention}. For a detailed understanding of the transformer network layer topology and model mechanics, refer to \cite{vaswani2017attention}. The changes introduced to the model to adapt it for SITS classification by \cite{Russwurm2020} include:
\begin{itemize}
    \item the addition of positional encoding to enable the model to utilise the sequential correlation of the time series
    \item transforming the time series with positional encoding into higher-level \emph{D}-dimensional feature representations through \emph{L} Transformer blocks. Transformer blocks are used to encode features through a multihead self-attention mechanism which is followed by multiple dense layers that are independently applied to each time instance
    \item the introduction of skip connections between the layers and layer normalisation across the entire model  
\end{itemize}

The last layer is then reduced through global maximum pooling along the temporal axis which is then projected to scores for each class by a final fully connected layer that applies a softmax activation function.

The configurable hyperparameters for this architecture include the number of attention heads, the number of transformer encoder layers, d\_model, d\_inner and the dropout rate. In \cite{Russwurm2020}, they use a d\_model of 512, a d\_inner of 2048, 6 transformer encoder layers, 8 attention heads and a dropout rate of 0.2. The model is compiled with an Adam optimiser using a learning rate and weight decay sampled from log-uniform distributions of $[10^{-8}, 10^{-1}$ and $[10^{-12}, 10^{-1}$ respectively. They do not state the optimally found values and thus are not reported here. This paper searches the above hyperparameters across the following spaces: the number of attention heads $H ~ ([1, 10])$, the number of transformer encoder layers $L ~ ([2, 20])$ in increments of 2, d\_model $M \in {2^5, 2^6, 2^7}$, d\_inner $I \in {2^6, 2^7, 2^8}$.

\section{Overview of study sites and experimental setup}
\label{DS}
This section provides an overview of the two datasets used in the training and evaluation of the studied methods. Summaries of the two datasets and their reference data are first given. This is then followed by an overview of the data preprocessing steps. Finally, the performance metrics are briefly discussed. 

\subsection{TiSeLaC dataset}
The first dataset comes from the Time Series Land Cover Classification Challenge (TiSeLaC) \cite{ienco_2017} that covers Reunion Island and is roughly 80MB in size. Between the training and test datasets, there are a total of 99,687 time series represented by pixels from 23 satellite images that are taken over the annual period of 2014 at a spatial resolution of 30m, and a long revisit cycle of 16 days \cite{qin2015mapping}. Each image contains 2866x2633 pixels but many pixels are left blank. These images use the L2A processing level, with the source data having been further processed to fill cloudy observations via a pixel-wise multi-temporal linear interpolation on each of the multi-spectral bands (OLI) independently. This enabled the computation of 3 complementary radiometric indices (NDVI, NDWI and BI). This processing provides a total of 10 features per pixel at each timestamp.

Each sample in the dataset is temporally ordered, meaning that features 1 to 10 represent the first timestamp and features 220 to 230 represent the last timestamp. The order of the features is also consistent, being composed of 7 surface reflectance values (Ultra Blue, Blue, Green, Red, NIR, SWIR1 and SWIR2) and 3 indices calculated from these surface reflectance values (NDVI, NDWI and BI).

\begin{figure}[ht]
\centering
\includegraphics[width=0.7\textwidth,height=\textheight,keepaspectratio]{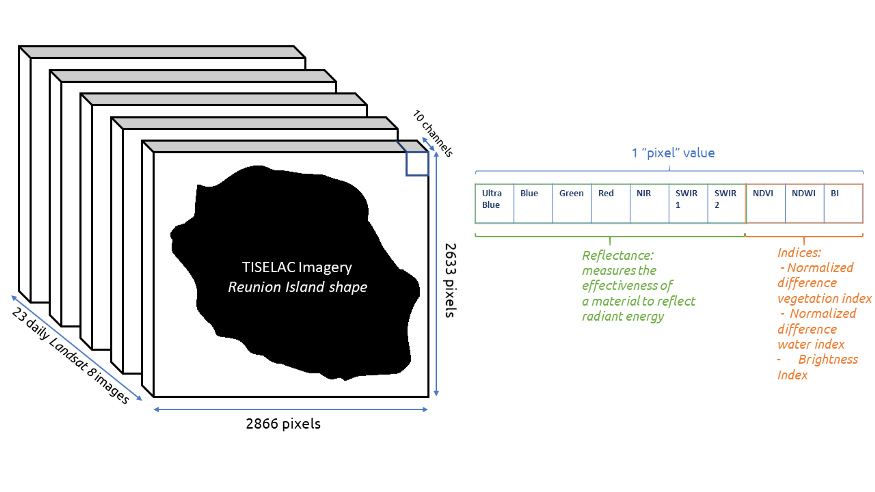}
\caption[TiSeLaC Study site]{The Reunion Island study site and the corresponding composition of the dataset. 7 channels are raw sensor values whereas the final 3 are calculated manually. Image sourced from \cite{martino_2020}.}
\label{fig:Study site of TiSeLaC dataset}
\end{figure}

Each sample is classified as one of 9 possible classes. The reference land-cover labels were gathered from two publicly available datasets, namely the 2012 Corine Land Cover (CLC) map and the 2014 farmers' graphical land parcel registration (RPG). Of the potential classes, the most significant ones were retained and spatial processing was applied which was aided by photo-interpretation. This was performed to ensure consistency with the image geometry. Pixel-based random sampling was applied to the dataset to produce the most balanced and representative ground truth possible. The distribution of classes in the TiSeLaC dataset is provided in Table~\ref{tab:TiSeLaC_Classes}. As can be seen, there are some noticeable class imbalances in the \emph{Other crops} and \emph{Other built-up surfaces} classes that the trained models will need to address. Figure~\ref{fig:Study site of TiSeLaC dataset} provides an overview of the TiSeLaC dataset composition.

\begin{table}[ht]
\centering
\begin{tabular}{ |c|c|c|c|  }
 \hline
\textbf{Class ID} & \textbf{Class Name} & \textbf{\# Instances Train} & \textbf{\# Instances Test} \\ \hline
1 & Urban Areas & 16000 & 4000 \\ 
2 & Other built-up surfaces & 3236 & 647 \\ 
3 & Forests & 16000 & 4000 \\ 
4 & Sparse Vegetation & 16000 & 3398  \\ 
5 & Rocks and bare soil	& 12942 & 2599 \\ 
6 & Grassland & 5681 & 1136 \\ 
7 & Sugarcane crops	& 7656 & 1531 \\ 
8 & Other crops & 1600 & 154 \\ 
9 & Water & 2599 & 519 \\
\hline
\end{tabular}
\caption[TiSeLaC class sample distribution]{\label{tab:TiSeLaC_Classes}Overview of class distributions in the training and testing datasets for the TiSeLaC dataset.}
\end{table}

\subsection{SITS-TSI dataset}
The second dataset named the SITS-TSI dataset originates from a paper by Tan et al. \cite{Tan2017}. The dataset consists of 46 chronological FORMOSAT-2 satellite images of an unspecified study area at an 8m spatial resolution with each image containing 1 million pixels, hence each image covers $64km^2$. Each pixel in the image forms a time series of a length of 46, resulting in a dataset of 1 million time series. The revisit time is not explicitly stated in \cite{Tan2017}, and neither is the time period over which the images are taken. However, other studies utilising FORMOSAT-2 data \cite{bsaibes2009albedo} note that FORMOSAT-2 imagery has a high temporal resolution of 1-3 days, improving upon that of the TiseLac dataset but does not provide a full annual reporting in this specific dataset. For each of the observations, only a single raster channel is provided per timestamp for each pixel, with there being no mention of the pre-processing used to acquire this value \cite{Tan2017}. The 8m spatial resolution would imply the images are multi-spectral being comprised of four 90 nm width wavebands \cite{bsaibes2009albedo}, but \cite{Tan2017} fail to mention any further processing that would result in each pixel containing only a single channel.

Each image has been manually corrected and labelled by geoscience experts using photo interpretation, urban databases and ground-sampling campaigns. This ensures a reliable and accurate ground-truth labelling for each sample which is classified as one of 24 possible classes. The class imbalances present within this dataset are more severe than those in the TiSeLaC dataset, and can be seen in Table \ref{tab:SITS-TSI_Classes} below. This dataset contains a number of classes with less than 10,000 instances in the training dataset, including classes: 4, 11, 13, 15, 16, 18, 20-24. It can be expected that these classes will represent a challenge to each of the classifiers.

The dataset provided by the authors \cite{Tan2017} was in the format of 10 cross-validation folds, with a 90:10 split between the training and testing data. Each fold contains 1 million samples which total 270MB in size. Details of how this cross-validation set was created are not included within \cite{Tan2017}, but in all cross-validation folds, each of the 24 classes are proportionally represented. It is worth mentioning that in order to identify possible spatial bias in the classification results when choosing any of the specific folds provided to train on, all 10 cross-validation data folds provided were individually trained and tested on, using the Temporal CNN as the comparative model (preserving the train/validation/test split proportions), and no significant changes in performance were observed. Therefore the selection of one fold over another is inconsequential as biases in the folds are similarly stochastic; confirming that the random sampling used is representative of the overall dataset. Hence, for the purposes of this paper, the first cross-validation fold is used, with the test set of this fold used to evaluate the performance of the trained models. The 90\% of data used for training is further split to an 80:20 ratio for training and validation sets which further improves the evaluation of each model. The number of samples in each dataset, therefore, becomes 720k,180k and 100k for the training, validation and test sets respectively. Figure~\ref{fig:Land cover map SITS-TSI} gives an example Land Cover map for the SITS-TSI dataset.

\begin{table}[ht!]
\centering
\begin{tabular}{ |c|c|c| }
 \hline
 \textbf{Class ID} & \textbf{\# Instances Train} & \textbf{\# Instances Test} \\
 \hline
 1 & 214641 & 23850 \\
 2 & 149424 & 16602 \\
 3 & 29361 & 3262 \\
 4 & 3899 & 434 \\
 5 & 33875 & 3764 \\
 6 & 84131 & 9347 \\
 7 & 43287 & 4810 \\
 8 & 58294 & 6477 \\
 9 & 135931 & 15104 \\
 10 & 11888 & 1321 \\
 11 & 946 & 105 \\
 12 & 22066 & 2452 \\
 13 & 1867 & 207 \\
 14 & 63106 & 7012 \\
 15 & 5736 & 637 \\
 16 & 514 & 57 \\
 17 & 16624 & 1847 \\
 18 & 3894 & 433 \\
 19 & 17374 & 1930 \\
 20 & 2206 & 245 \\
 21 & 177 & 20 \\
 22 & 282 & 32\\
 23 & 382 & 42 \\
 24 & 95 & 10 \\
 \hline
\end{tabular}
\caption[SITS-TSI class sample distribution]{\label{tab:SITS-TSI_Classes}Overview of class distributions in the training and testing datasets for the SITS-TSI dataset. }
\end{table}

\subsection{Dataset preprocessing}
To comprehensively evaluate the effectiveness of Temporal CNNs, the two datasets introduced earlier are used for training and testing purposes. Since the characteristics of these datasets vary greatly, they were used to evaluate how the following factors affect performance: the number and type of input feature channels, class imbalances, the number of output classes, temporal sequence length and the number of training samples available. Once conclusions were made in regard to these factors, further experiments investigated hyperparameter tuning and the altering of the architectural configuration. Both datasets \cite{ienco_2017, Tan2017} have been corrected geometrically and radiometrically to ensure that 1) each pixel (x,y) always maps to the same area, and 2) that the spectral information is consistent from one image to the next in the series. 

Prior to training, each dataset was split into a training, validation and test set. The test set size is dependent on the dataset being used, but the training data is always split to a ratio of 80:20 to provide training and test data. This was done to ensure that over-fitting could be monitored and a fair analysis could be made. The feature data in each dataset was also normalised using a min-max normalisation per feature type. Traditional min-max normalisation subtracts the minimum and then divides by the range (\emph{i.e.} the maximum value minus the minimum value). In \cite{Pelletier2019}, they note that this method is sensitive to extreme values and so instead use the 98\% percentile rather than the minimum or maximum value. For each feature type per timestamp, the percentile values are extracted and the normalisation is applied, assisting the back-propagation of data through the neural network. Normalisation ensures each feature value in the dataset is scaled between 0 and 1, ensuring each feature uses the same scale and distribution, ensuring no single input channel influences the model performance disproportionately. The target values were also converted into a one-hot encoding for use in the training step for calculating the loss.

\subsection{Performance metrics}
When evaluating the various classification algorithms, quantitative assessments are enabled via the use of the Overall Accuracy (OA) and F1 score. A per-class analysis is also provided to supplement the averaged results. The training time of each model was also investigated to help distinguish similarly performing models by their computational efficiency. This combination of evaluation metrics is commonly used for classification tasks and offers a diverse interpretation of the results beyond a basic assessment of the overall accuracy \cite{hossin2015review}. To monitor the progress of learning during the model's during training, Categorical Cross Entropy was employed \cite{zhang2018generalized}. Finally, to monitor overfitting, training graphs and cross-validation techniques were employed to assure confidence in the training and testing of each model across all classes.

\section{Experimental Results}
\label{exp}

This section reports the results of the experiments carried out on the methods discussed in Section \ref{surveyed methods}. These experiments are designed to answer the following questions:

\begin{itemize}

    \item Which alterations to the Temporal CNN architectural configuration has the most noted effect on performance; e.g. the width, depth and filter size used by the model?
    \item Which hyperparameter values for the employed regularisation methods provide optimal performance on the Temporal CNN, and whether these values are particularly sensitive to the dataset in use?
    \item Do the performances and training times of the surveyed methods for SITS classification agree with prior observed results in the literature?
    \item Which dataset-dependent features have the most notable effect on performance across the surveyed architectures?
    
\end{itemize}

Due to processing limitations, only the best experimental result for each model's performance and time is reported in place of an average of multiple runs. To enable a fair comparison of training times for each methodology, when comparing times, each implemented neural network was trained using a batch size of 128 for 20 epochs without early stopping. When searching for the best performance on each model, training was conducted until the early stopping mechanism that monitored the validation loss ended training, ensuring each model has reached its full potential before noticeable overfitting occurred.

To clearly answer the questions the experiments pose, the rest of this section is broken down into two core sections: Section \ref{TCNN_Experiments} and Section \ref{Survey_Results}. Section \ref{TCNN_Experiments} explores the various investigations into the Temporal CNN architecture. This covers the discovery of the optimal architectural configuration for the Temporal CNN on both studied datasets using recommendations from Pelletier et al. \cite{Pelletier2019}, and an overview of the investigation into regularisation for the Temporal CNN and the final optimised hyperparameter values. Section \ref{Survey_Results} begins by providing the optimised hyperparameter values for each of the surveyed architectures that were introduced in Section \ref{surveyed methods}. The bulk of this section reports the performances of each benchmark model against the Temporal CNN for the TiSeLaC and SITS-TSI datasets respectively. Further experimental results conducted on the two datasets for the Temporal CNN architecture that aren't crucial to this discussion are provided in Appendix \ref{secA2}. These additional results document the variations in architectural width, depth, filter size and degree of regularisation used.

\subsection{Experiments on the Temporal CNN architecture}
\label{TCNN_Experiments}
Testing of the Temporal CNN architecture included experiments that analysed factors such as how big and deep to construct models, the effect of changing the kernel size and how to control over-fitting by adjusting the dropout and batch size. Due to the smaller dataset size, experiments were carried out on the TiSeLaC dataset utilising the reduced training time. Promising architectural configurations were then applied to the SITS-TSI dataset to confirm the ability of the model to generalise.

At the centre of this investigation is the prevalence of the \emph{bias-variance-trade-off}. Since dataset size is a key component of the bias-variance-trade-off, the optimal model settings found for each dataset may vary.  Within machine learning, the more complex a model is (\emph{i.e.} more parameters), the higher its bias. Lower bias translates to a lower number of errors when making predictions on the training data, which may reduce the generalisation of the model. On the other hand, an increase in variance is observed as the model complexity increases for a fixed number of training samples. If the complexity of a model outstrips the provided data for training, it cannot effectively learn to accommodate new data outside of the observed training distribution \cite{Pelletier2019}.

Controlling this trade-off between bias and variance within neural networks requires careful consideration of model complexity and supplementary regularisation techniques. Within neural networks, the bias is solely influenced by the complexity of the model, whereas the variance is dependent on the amount of training data provided. Similar experimentation on model complexity conducted within \cite{Pelletier2019} uses the number of parameters within each generated model as a proxy for model complexity as their study context had a static data size and the number of classes. This paper investigates model performance on two datasets, each being of vastly different sizes and containing a different number of classes. Therefore, the number of parameters cannot be as so assuredly linked to an increase in complexity. Model complexity within this context is hence observed as a combination of the depth (\emph{i.e.} number of convolutional layers) and width (\emph{i.e.} the number of units in the convolutional and dense layers). Investigations into the optimal depth range from using 2 convolutional layers up to 5, with values for the number of units within the convolutional and dense layers ranging from 64 to 2048.

\subsubsection{Investigating Optimal Width}
To investigate the optimal width of each model, 50 architectures were generated by varying the configuration of various parameters. Of these 50 architectures, only the results with consistent values for the dropout, batch size and filter size are studied. More extreme configurations were tested on the TiSeLaC dataset due to the smaller size of the dataset. Promising configurations from these experiments were then applied to the SITS-TSI dataset and then further experimented on to test for differences in optimisation settings between the two datasets.

To find the optimal width of the network, the number of units in the convolutional layers was varied alongside the number of units in the fully connected layer. This method of experimentation generated insights into the number of required units for effective feature extraction in the convolutional layers, as well as what ratio of units to use in the subsequent fully connected dense layer. In accordance with the findings of \cite{Pelletier2019} and \cite{russwurm2019breizhcrops} the number of convolutional units used is never below 64 due to adverse effects on performance. The results for these experiments are reported, with each model comprised of three convolutional layers, a dense layer and a Softmax layer. Dropout is set to 0.2, the filter size is set to 5 and a batch size of 128 is employed. The results in Table~\ref{tab:Width_Results_TiSeLaC} and Table~\ref{tab:Width_Results_SITS-TSI} display the accuracies of the various models chosen with the bold accuracy of each table indicating the best-performing architecture.

\begin{table}[ht!]
\centering
\begin{tabular}{ c c c  }
 \hline
 \textbf{\# Convolutional Units} & \textbf{\# Fully Connected Units} & \textbf{OA} \\
 \hline
 64 & 64 & 90.9\\
 \hline
 64 & 128 & 92.64\\
 \hline
 64 & 256 & 92.81\\
 \hline
 64 & 512 & 91.33\\
 \hline
 128 & 128 & 93.57\\
 \hline
 128 & 256 & \textbf{95.02}\\
 \hline
 256 & 256 & 94.58\\
 \hline
 512 & 256 & 93.48\\
 \hline
 512 & 512 & 93.36\\
 \hline
 512 & 1024 & 93.02\\
 \hline
 512 & 2048 & 93.54\\
 \hline
 1024 & 256 & 93.83\\
 \hline
\end{tabular}
\caption[Temporal CNN width experiments TiSeLaC]{\label{tab:Width_Results_TiSeLaC}Results for adjusting the width of the model on the TiSeLaC dataset.}
\end{table}
 
\begin{table}[ht!]
\centering
\begin{tabular}{ c c c  }
 \hline
 \textbf{\# Convolutional Units} & \textbf{\# Fully Connected Units} & \textbf{OA} \\
 \hline
 64 & 256 & 86.95\\
 \hline
 128 & 256 & \textbf{87.16}\\
 \hline
 256 & 128 & 86.87\\
 \hline
 256 & 256 & 87.13\\
\end{tabular}
\caption[Temporal CNN width experiments SITS-TSI]{\label{tab:Width_Results_SITS-TSI}Results for adjusting the width of the model on the SITS-TSI dataset.}
\end{table}

Although the tested configurations are not exhaustive, the results confer a number of conclusions. For the SITS-TSI dataset, since it only uses a single feature per timestamp, adjusting the width of the network seemed to have almost no effect on the observed performance as the extra neurons could not be effectively utilised. For both datasets, performance decreases when the number of convolutional units is greater than the number of fully connected units. For the TiSeLaC dataset, optimal results are displayed when there are either 128 or 256 units in the convolutional layers, with the number of fully connected units being either equal to or double the number of convolutional units. Smaller and larger numbers of units for the convolutional layers display adverse performance. The model either lacks the computational complexity to extract relevant features or has too great a complexity that it loses the ability to sufficiently generalise. Experiments on the SITS-TSI dataset that used 256 convolutional units and 256 fully connected units produced the greatest accuracies.

Given these observations, the recommendation would be to employ either 128 or 256 units in the convolutional layers and 256 in the fully connected layer. Using the greater number for the choice of convolutional units will increase model complexity, resulting in longer training times and an increased risk of over-fitting. Adjusting the batch size and dropout therefore must be included during model development.

\subsubsection{Investigating Optimal Depth}
To investigate the depth of the neural network, 8 architectures were created: 4 per studied dataset. The number of convolutional layers varies between 2 and 5. Results from \cite{Pelletier2019} demonstrated markedly reduced performance for values outside of the previously stated range. Their experiments kept the model complexity the same by reducing the width of the models as the number of layers increased. For the experiments within this paper, since the model complexity is partly dataset dependent, the complexity cannot be controlled in the same manner. Hence, the width of the architectures is kept constant in order to make fair comparisons. The width of each model is kept at 128 convolutional units and 256 fully connected units in accordance with the findings of the previous section. These experiments contained various widths and dropout values. It was found that as the complexity of the model increased, updating the dropout correspondingly improved performance as expected. 

\begin{figure}[ht]
\centering
\includegraphics[width=0.7\textwidth,keepaspectratio]{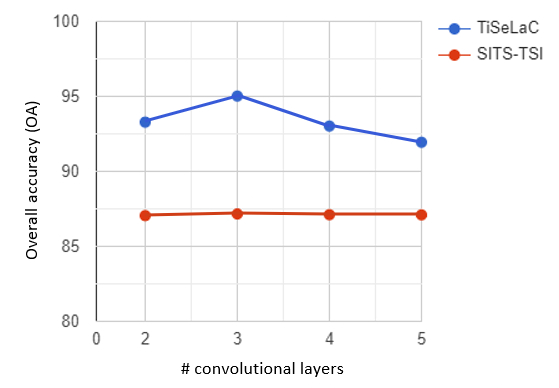}
\caption[Effect of depth on performance]{Overall accuracy as a function of the number of convolutional layers (depth influence) on the TiSeLaC and SITS-TSI datasets.}
\label{fig:Depth_Results}
\end{figure}

Figure~\ref{fig:Depth_Results} shows that for the TiSeLaC dataset, 3 convolutional layers are optimal, whereas there is no discernible difference for the SITS-TSI dataset. It can be concluded the usage of 3 convolutional layers is optimal for both datasets. Values lower than this may reduce the ability of the model to learn effective feature representations, and more may increase complexity to the degree that it reduces generalisation and training speed. If a greater number of layers is desired, then an increase in batch size and dropout is recommended to combat over-fitting.

Due to the choice of the dataset, the number of convolutional layers, and the width of the network, the number of parameters in each constructed model range from 140,000 to $\sim$18 million. The optimal model for the TiSeLaC dataset contained $\sim$2.1 million parameters and $\sim$3.8 million for the SITS-TSI dataset. The increase in parameters for the SITS-TSI dataset results from the number of output classes being predicted for; 24 instead of 9. In general, the number of parameters does not significantly affect the performance of the model, with results ranging from $\sim$91-95\% for the TiSeLaC dataset, and 87 $\pm$ 0.25\% for SITS-TSI. Using a model with a lower number of parameters is preferable since the complexity and training time can be dramatically reduced. 

If models need to be developed for a fixed complexity (\emph{i.e.} the number of parameters) given specific hardware or software limitations, the use of an inappropriate number of convolutional layers or the number of units in the convolutional and fully-connected layers may lead to an underestimated Temporal CNN. In situations where additional resources are available, more exhaustive and computationally expensive cross-validation procedures could be used to further optimise the Temporal CNN architecture and avoid developing under-optimised models. Overall, through these experiments, the optimal architecture for the two datasets was found to be very similar, with the only major difference being the optimal kernel size, which is justified in the following section.

\subsubsection{Influence of filter size}
CNNs that employ temporal guidance perform convolutions across the temporal domain of a sample time series. Investigating the size of the filter for these convolutions is of special interest for finding which filter size works best for the temporal sampling used by each studied dataset. The gap in temporal sampling for the two datasets is not explicitly reported so it is unknown how a filter size of $f$ will abstract the temporal information. The filter $f$ (f being odd) abstracts the temporal information over $\pm (f-1)$ timestamps, before and after each point in the time series. This is commonly coined the \emph{reach} of the convolution, corresponding to half of the width of the temporal neighbourhood used for the temporal convolutions \cite{Pelletier2019}. Note that this definition can only be used in circumstances where the time series used is regularly sampled, with the reach depending on the number of days that separate acquisitions. Hence, for this reason, this paper will explore experiments in relation to the filter size and not reach as a regular temporal sampling in both datasets is not observed. Experiments conducted on the temporal reach in \cite{Pelletier2019} found large filter sizes reduced the quality of the temporal resolution within the SITS data. Their studied dataset had considerably more timestamps, and also a denser temporal frequency that allowed them to experiment with larger filter sizes. This investigation hence focused on a smaller number of filter sizes, investigating three filters $f = {3,5,7}$ for the TiSeLaC dataset as the temporal length is only 23 timestamps. Four filters are investigated for the SITS-TSI dataset where $f = {3,5,7,9}$. Since the temporal length is 46, an increase in the upper filter bound used could be investigated. Models with the following configuration were used during the selection stage of the filter size: 128 units per convolutional layer, 256 units in the fully-connected layer, a batch size of 128 and a dropout of 0.2.

\begin{table}[ht!]
\centering
\begin{tabular}{ r c c c c }
 \hline
 \textbf{Filter Size} & 3 & 5 & 7 & 9\\
 \hline
 \textbf{TiSeLaC OA} & 93.94 & \textbf{95.02} & 93.93 & NA \\
 \hline
 \textbf{SITS-TSI OA)} & 86.70 & 86.80  & \textbf{86.98} & 86.91 \\
 \hline
\end{tabular}
\caption[Temporal filter experiments]{\label{tab:Filter_Size_Results}Overall accuracy of each model as the filter size is fine-tuned. The bold value indicates the highest result.}
\end{table}

Table~\ref{tab:Filter_Size_Results} highlights that for the TiSeLaC dataset the maximum OA is achieved with a filter size of 5 whereas, for the SITS-TSI dataset, it is achieved with a filter size of 7. One factor that may have caused this result is the greater number of temporal observations in the SITS-TSI dataset. Results seem to be relatively consistent for the SITS-TSI dataset, however, so these findings are not alarming. The effect of the temporal filter size is slightly more pronounced in the TiSeLaC dataset. Generally, however, these results demonstrate the importance of high temporal resolution within SITS \cite{Pelletier2019}. Due to the Sentinel-2 and FORMASAT-2 satellites having high acquisition frequencies, the Temporal CNN is able to abstract enough temporal information from the convolution operations. Crucially, the optimal value for the filter size used heavily depends upon the focus of the SITS data. Tasks such as crop monitoring and urban expansion require much larger temporal filter size values when capturing useful information for classification.

\subsubsection{Controlling over-fitting}
\label{parameters}

Similar studies have investigated regularisation techniques for Temporal CNNs \cite{Pelletier2019}, focusing on the use of four regularisation techniques: 1) regularisation of the weights, 2) dropout, 3) usage of a validation set and 4) using batch normalisation layers. The conclusions of \cite{Pelletier2019} were that dropout had the most significant effect, whilst weight decay and the use of a validation test were nominal. Although insignificant for the goal of improving performance, the validation set is helpful for monitoring over-fitting and as such is retained here. The experiments conducted for controlling over-fitting, therefore, investigate the optimal value to use for dropout for each studied dataset.

\subsubsection*{Selecting the batch size}
The batch size was found to have no notable adverse effects on performance, as was reported by Pelletier et al. \cite{Pelletier2019}. Increasing the batch size  does however cause a significant reduction in training time, whereby a doubling in the batch size corresponds to roughly halving the training time. The recommendation is to therefore use larger values for the batch size if memory is available as training can be vastly sped up, allowing for a greater degree of experimentation. A batch size of 128 is generally used to train Temporal CNN models on the TiSeLaC dataset, and 256 for the SITS-TSI dataset. When running time comparisons between models, however, the batch size is consistently kept at 128 for fairness.

\subsubsection*{Selecting a value for dropout}
Dropout is a technique for addressing overfitting \cite{srivastava2014dropout}, being the prime mechanism for ensuring generalisation within trained models as reported by \cite{Pelletier2019}. The idea is to temporarily drop random units and their connections from layers within the neural network during training, preventing units from co-adapting too much. This process is done every epoch, and perhaps per mini-batch, thus creating an exponential number of 'thinned' models. The presence of neurons is made unreliable, forcing the model to generalise through the sampling of many thinned networks. A parameter \textbf{\emph{p}} is used to denote the percentage of neurons in a layer that is randomly dropped. 

\begin{figure}[ht]
\centering
\includegraphics[width=0.6\textwidth,height=\textheight,keepaspectratio]{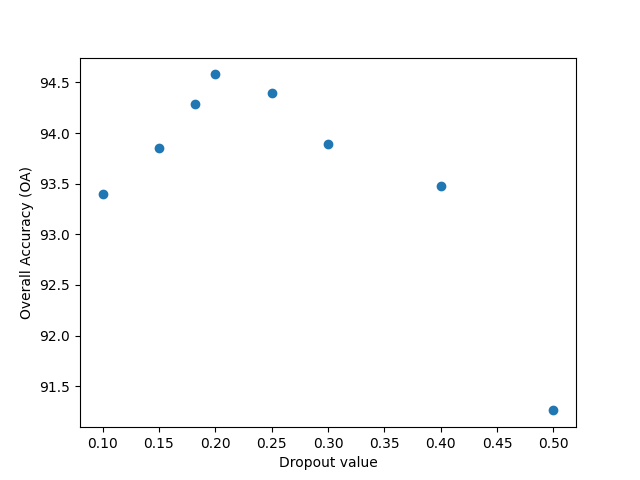}
\caption[OA versus Dropout]{Overall Accuracy of each model as a dropout is fine-tuned. Significant performance drops are witnessed at higher values.}
\label{fig:Dropout_vs_OA}
\end{figure}

Dropout values between 0.1 and 0.5 were tested on the TiSeLaC data, with the results influencing the range of tested values on the SITS-TSI dataset. Larger values for dropout ($>0.3$) reduced the accuracy of models due to too much information loss, whereas lower dropouts ($<0.15$) did not account for over-fitting enough. From Figure~\ref{fig:Dropout_vs_OA}, it can be shown that values between 0.15 and 0.25 are optimal due to their comparable values. The specific value of 0.182 was experimented with due to it being used as the dropout value for the pre-trained Breizhcrops Temporal CNN \cite{russwurm2019breizhcrops}, which itself was subsequently found by using a random sampling strategy for 12 hours on a GPU. Following the results of this experiment, a dropout of 0.2 is adapted as the default value for future use and for the SITS-TSI dataset. \\ 

\begin{figure}[ht]
\centering
\includegraphics[width=\textwidth,height=\textheight,keepaspectratio]{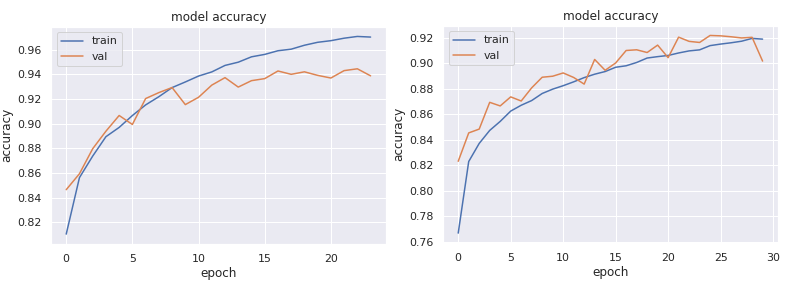}
\caption[Effect of dropout on over-fitting]{Training curve of the model with 0.2 dropout (L) against the training curve of the model with 0.5 dropout (R).}
\label{fig:Dropout}
\end{figure}

As can be seen in Figure~\ref{fig:Dropout}, the over-fitting appears to be more controlled with a dropout of 0.5, but due to higher information loss the validation accuracy by 3.2\%. The over-fitting observed with a dropout of 0.2 is not too severe and training can be stopped earlier.

\subsection{Experimental results for surveyed methods}
\label{Survey_Results}

This section presents the experimental results for the surveyed methods across the two studied datasets, including a discussion and analysis of the findings. 

\subsubsection{Optimal architecture configuration and hyperparameter selection for comparative models}
\label{optimal_settings}

For the model selection, models were trained in turn on the training sets of each dataset, with the validation loss being used to determine if the chosen configuration and hyperparameters resulted in a more optimal model. The optimal model was then evaluated on the test data for each dataset as the basis for the final review. Each model is trained with enough epochs to allow the early stopping mechanism to terminate training; balancing performance against overfitting.

The optimal hyperparameter values as discussed in Section \ref{surveyed methods} are given here. For the Time-CNN architecture, the optimal dropout was found at 0.15 for the TiseLaC dataset and 0.3 for the SITS-TSI dataset. Both datasets have $\nu$ = 0.001 and $\lambda$ = 0.0, selected from \cite{IsmailFawaz2019} for both datasets. The MCDCNN had a dropout of 0.182 for the TiseLaC dataset and 0.4 for the SITS-TSI dataset. Both datasets use $\nu$ = 0.01 and $\lambda$ = 0.0005, also from \cite{IsmailFawaz2019}. The optimal InceptionTime models had $L = 4$ and $H = 128$ for the TiSeLaC dataset and $L = 6$ and $H = 256$ for the SITS-TSI dataset. The values for $\nu$ and $\lambda$ are an approximation of the settings found in \cite{russwurm2019breizhcrops}, with the values 0.01 and $2 \cdot 10^{-6}$ respectively for each dataset. Regarding the RNN, the optimal TiSeLaC model had $L = 3$, $H = 256$, $F = 512$, and a dropout rate of 0.25. For the SITS-TSI dataset, the optimal settings were $L = 3$, $H = 128$, $F = 256$, and a dropout of 0.2. In Pelletier et al. \cite{Pelletier2019} where the RNN model is from, they use $\nu = 1 \cdot 10^{-6}$ and a default $\lambda = 0$ which is applied to both datasets here. Lastly, the Transformer model was found to share the same optimal parameters for both datasets with $H = 2$, $L = 10$, $M = 128$, $I = 256$, and dropout as 0.03. As from recommendations from \cite{russwurm2019breizhcrops}, $nu = 1.31 \cdot 10^{-3}$ and $lamda = 5.52 \cdot 10^{-8}$.

\subsubsection{Overview of results for the TiSeLaC dataset}
\label{ds1}

The performance of each optimised model is given in Table~\ref{tab:TiSeLaC_Results_Acc}. The reported accuracies were produced from predictions on the test set. Despite extensive hyperparameter optimisations and the employment of procedures to generalise the model, some over-fitting was still observed on the InceptionTime network due to its complexity. The observed test accuracies do however demonstrate the ability of each model to generalise well to unseen data. The training times to train each model to 20 epochs without early stopping and a batch size of 128 are reported in Table~\ref{tab:TiSeLaC_Results_Time}. 

\begin{table}[ht!]
\centering
\begin{tabular}{ |l|c|c|}
\hline
\multicolumn{1}{|c|}{\textbf{Method - TiSeLaC dataset}} & \multicolumn{1}{|c|}{\textbf{OA}} & \multicolumn{1}{|c|}{\textbf{F1}}\\
\hline
Temporal CNN & \textbf{95.0} & \textbf{94.9} \\
\hline
Time-CNN & 89.2 & 89.1 \\
\hline
MCDCNN & 88.2 & 88.3 \\
\hline
InceptionTime & 92.9 & 93.0 \\
\hline
RNN & 92.7 & 92.6 \\
\hline
Transformer & 91.6 & 91.7 \\
\hline
\end{tabular}
\caption[TiSeLaC performance results]{\label{tab:TiSeLaC_Results_Acc}Best accuracies on the TiSeLaC dataset for each method.}
\end{table}

\begin{table}[ht!]
\centering
\begin{tabular}{ |l|c|c|}
 \hline
 \multicolumn{1}{|c|}{\textbf{Method - SITS-TSI dataset}} & \multicolumn{1}{|c|}{\textbf{Time (h)}}\\
 \hline
 Temporal CNN & 125 \\
 \hline
 Time-CNN & \textbf{60} \\
 \hline
 MCDCNN & 73 \\
 \hline
 InceptionTime & 410 \\
 \hline
 RNN & 275 \\
 \hline
 Transformer & 383 \\
 \hline
\end{tabular}
\caption[TiSeLaC training time results]{\label{tab:TiSeLaC_Results_Time}Best training times for each method on the TiSeLaC dataset using similar hyperparameter settings.}
\end{table}

As evidenced from table \ref{tab:TiSeLaC_Results_Time}, the Temporal CNN outperformed the other approaches by at least $\sim$2\%. The Time-CNN outperformed the MCDCNN as expected due to it being reported as an improved version of the MCDCNN. Despite being a simpler model than the Time-CNN, the MCDCNN was slower in this case since it had to separate each spectral channel for individual processing and then recombine them, unlike the Time-CNN with performs these operations simultaneously. The overall simplicity of the MCDCNN architecture displays a respectable accuracy but it is outclassed by every other method as it lacks the complexity to extract as meaningful feature maps as the other classifiers. A drawback evidenced in the lower performance of the MCDCNN is that the features in the TiSeLaC dataset are not inherently independent of each other as the last three features represent calculated indices that are derived from values in the prior seven features
\cite{ienco_2017}. Furthermore, depending on the target associated with a feature vector, many of the channel values may be strongly correlated with each other, reducing the basis of the argument for the independence of features. Given that the core of the methodology is the use of a generic CNN with only minor modifications to process temporal data, it is unsurprising to observe it achieving the lowest performance. By utilising temporal and spectral guidance, the Temporal CNN was able to discern the differences between each class in relation to their temporal and spectral evolution. This led the classifier to generate the most optimal feature maps that extend to accurate predictions.

Whilst the speed of the Temporal CNN is modest compared to that of the Time-CNN and MCDCNN, it has less than half the computational cost of the next fastest competitor: the RNN. Beyond the Temporal CNN there is a significant jump in training times, with each model gaining ~2-3\% accuracy over the Time-CNN but having training times at least 4.5 times longer, making the trade-off between performance and training cost questionable. Overall in this context, the training time as a performance metric is a lesser concern due to the small dataset size and the comparatively low training times observed, but it does serve as an indicator for performance on larger datasets.

Despite the InceptionTime, RNN, and Transformer models all achieving respectable accuracies, the comparatively long training times are indicative of a poor ability to scale to large data domains. Although these models boasted strong performances, in this instance, they were outperformed by the relatively simpler Temporal CNN in terms of speed and accuracy. It could be the case that these networks, especially the complex InceptionTime network perform better within larger data domains since the complexity of the networks would enable them to effectively exploit larger numbers of training samples. As observed, in the above results, this would substantially increase training times, potentially to a prohibitive degree. Given that each of the models was trained on the same parallel GPU hardware, the training times for each model are solely linked to the number of parameters and the architectural dynamics of each model. Transformers have previously been cited as being faster than convolutional and recurrent-based models \cite{vaswani2017attention, SainteFareGarnot2020}, but in the case of this dataset, the Transformer was the second slowest model. In the case of prior studies that compare training times in relation to the number of model parameters \cite{russwurm2019breizhcrops}, Temporal CNNs and LSTMs can have significantly more parameters than a Transformer model and still be trained faster, due in this case to both models needing fewer layers and computational units than the Transformer. The optimised Transformer required 10 Transformer encoder layers which added computational complexity, inflating the training time considerably. InceptionTime is even slower than the Transformer due to the optimal model requiring 256 hidden units within each layer, increasing the number of parameters significantly.

\begin{table}[ht!]
\centering
\begin{tabular}{ |l||c|c|c|c|c|c|c|  }
 \hline
 & \multicolumn{6}{c|}{\textbf{Per class accuracies for each model - TiSeLaC dataset}} \\
 \hline
 Class & Temporal CNN & Time-CNN & MCDCNN & InceptionTime & RNN & Transformer\\
 \hline
 1 & 93.50 & 90.78 & 93.13 & \textbf{95.55} & 94.05 & 94.37\\
 \hline
 2 & \textbf{86.94} & 65.85 & 58.42 & 64.43 & 76.66 & 76.10\\
 \hline
 3 & \textbf{94.89} & 91.27 & 88.38 & 93.20 & 92.54 & 92.74\\
 \hline
 4 & 95.93 & 91.66 & 90.12 & \textbf{96.37} & 95.31 & 93.14\\
 \hline
 5 & \textbf{96.68} & 95.23 & 91.11 & 94.97 & 95.02 & 92.93\\
 \hline
 6 & 90.77 & 82.27 & 76.62 & \textbf{91.67} & 86.39 & 88.07\\
 \hline
 7 & \textbf{97.47} & 94.66 & 95.12 & 96.55 & 95.77 & 93.85\\
 \hline
 8 & \textbf{66.67} & 37.63 & 35.51 & 57.38 & 44.44 & 42.92\\
 \hline
 9 & \textbf{92.35} & 85.49 & 88.40 & 89.62 & 91.00 & 88.91\\
 \hline
\end{tabular}
\caption[TiSeLaC per-class performance results]{\label{tab:TiSeLaC_Per_Class_Accs}Best accuracies on the TiSeLac dataset for each method, per class.}
\end{table}

By comparing the predictive power of the Temporal CNN against the weakest model (MCDCNN), comparisons can be made on how each model handles under-sampled classes and classes with similar spectral and temporal features. By looking at Table \ref{tab:TiSeLaC_Per_Class_Accs}, it can be shown that both models are affected by the low frequency of the \emph{other crops} (8) class, which both models frequently misclassify it as \emph{forest}. The MCDCNN also frequently misclassifies \emph{other built-up surfaces} (2) as \emph{urban areas}(1) due to their similar properties. The overall performance of the Temporal CNN was damaged by the sub-par performance on the under-sampled \emph{other crops} (8) class which was correctly classified 66.7\% of the time compared to an average of 90\%+ for the other classes. On a per-class basis, either the Time-CNN or MCDCNN had the weakest performances except for the Transformer on \emph{sugarcane crops} (7). The Temporal CNN was the stronger predictor for 6 of the classes, with InceptionTime being the strongest for the other 3. Remarkably, the Temporal CNN was able to outperform all of the other classifiers by a noticeable margin for the under-sampled classes (2, 6, 8, 9), highlighting its ability to make accurate predictions with few training samples. All models shared a similar weakness in classes that had similar spectral and temporal profiles. Other built-up areas (2) are an under-sampled class and have a similar spectral and temporal to Urban areas (1), resulting in frequent misclassifications. It was also observed that Other crops (8) which is the least represented class, had the lowest accuracy on every model, commonly misclassified as Forests (3), Sugarcane crops (7) and Urban areas (1). This is a strong indication that despite the strength of any individual model, performance is severely affected in classes with few training samples available.

It can be demonstrated that the Temporal CNN works well when there are many features available per data point, with the model seemingly able to learn information encodings within the temporal dimension of the data within a short training time. The performance of the other methods is also respectable. The simpler CNN models of the Time-CNN and MCDCNN proposed for the dataset were both outperformed by the more sophisticated architectures (\emph{i.e.} Temporal CNN, RNN, InceptionTime and Transformer) by at least 2\%. In SITS datasets that contain a multitude of input channels in which labelled data are less abundant, models that can fully leverage the input channels across the temporal domain generally yield better results. Of these specialised architectures, the Temporal CNN performed best with the contemporary RNN architecture narrowly outperforming the next best CNN architecture. The GRU layers of the RNN were able to effectively extract relevant features from the temporal dimension of the dataset. The InceptionTime architecture was also able to do this to a similar degree, but the complexity of the model outstripped the number of available training samples. For a smaller number of training samples, the RNN could learn effective feature representations more adeptly than the InceptionTime architecture, hinting that for the composition of the TiSeLaC dataset, the RNN architecture is more appropriate. The long training time of the RNN does however indicate a potential problem for scalability, especially given its relevant simplicity compared to the InceptionTime network. It was surprising to see the RNN outperform the Transformer network given that Transformers have previously surpassed the performance of RNNs, partly attributed to Transformers not suffering the vanishing gradient problem since they substitute recurrent layers for self-attention and positional encoding. The likely explanation is that the time-series length for this dataset is not long enough for the vanishing gradient to degrade performance to a noticeable degree. 

As reflected in these results, the optimised Temporal CNN architecture was able to outperform both the specialised and simpler architectures alike. Of the specialised architectures, it also had the shortest training time by a considerable margin: demonstrating that models with high complexity are not essential for state-of-the-art results. Many of the experimental Temporal CNN architectures investigated in the experiments section and Appendix \ref{secA2} routinely outperformed all of the other optimised models, demonstrating that extensive hyperparameter tuning is not a necessity for outstanding performance. These results demonstrate the intrinsic advantages Temporal CNNs possess for efficiently processing multi-spectral SITS data. 

\subsubsection{Overview of results for the SITS-TSI dataset}
\label{ds2}

The Temporal CNN model studied in section \ref{ds1} was adjusted and optimised for use on the SITS-TSI dataset which was then compared to the five other models also optimised on the dataset. Prior to analysing the following results, a technical note needs to be raised. Since the SITS-TSI dataset only has one feature channel per timestamp, the MCDCNN essentially acts as a single channel 1D-CNN. This simple model was expected to be outperformed by the other neural networks but the results are still shown to be respectable. Tables~\ref{tab:SITS-TSI_Results_Acc} and ~\ref{tab:SITS-TSI_Results_Time} provide the accuracies and training times of each model for the SITS-TSI dataset respectively.

\begin{table}[ht!]
\centering
\begin{tabular}{ |l|c|c|}
\hline
\multicolumn{1}{|c|}{\textbf{Method - SITS-TSI dataset}} & \multicolumn{1}{|c|}{\textbf{OA}} & \multicolumn{1}{|c|}{\textbf{F1}}\\
\hline
Temporal CNN & \textbf{87.3} & \textbf{87.2} \\
\hline
Time-CNN & 86.5 & 86.4 \\
\hline
MCDCNN & 85.8 & 85.7 \\
\hline
InceptionTime & 87.1 & 87.0 \\
\hline
RNN & 85.5 & 85.6 \\
\hline
Transformer & 85.9 & 85.9 \\
\hline
\end{tabular}
\caption[SITS-TSI performance results]{\label{tab:SITS-TSI_Results_Acc}Best accuracies on the SITS-TSI dataset for each method.}
\end{table}

\begin{table}[ht!]
\centering
\begin{tabular}{ |l|c|c|}
 \hline
 \multicolumn{1}{|c|}{\textbf{Method - SITS-TSI dataset}} & \multicolumn{1}{|c|}{\textbf{Time (h)}}\\
 \hline
 Temporal CNN & 0.181 \\
 \hline
 Time-CNN & 0.124 \\
 \hline
 MCDCNN & \textbf{0.0976} \\
 \hline
 InceptionTime & 4.34 \\
 \hline
 RNN & 0.857 \\
 \hline
 Transformer & 1.44 \\
 \hline
\end{tabular}
\caption[SITS-TSI training time results]{\label{tab:SITS-TSI_Results_Time} Training times for each method on the SITS-TSI dataset using a batch size of 128 for 20 epochs without early stopping.}
\end{table}

The results on the SITS-TSI dataset highlight that of the models tested, CNN approaches largely yielded superior performances than the RNN and Transformer. With less than a 2\% difference in accuracy between the strongest (Temporal CNN) and weakest (RNN) models, each of the architectures was able to demonstrate their ability to learn robust feature extractors from the dataset. These features were then able to accurately inform the class predictions of each network, allowing them to make strong predictions. 

Of the tested models, the Temporal CNN displayed the best performance with an accuracy of 87.3\% over the 24 class labels, 0.2\% higher than the next best-performing model being InceptionTime. Although the InceptionTime model was a close competitor to the Temporal CNN, it was by far the slowest of the neural networks, with the Temporal CNN significantly outpacing it. When investigating training times, the MCDCNN was the fastest network due to the relatively low number of parameters it contains and the comparatively simple operations it performs. The Time-CNN in this instance was not the fastest since the MCDCNN did not need to separate out each feature channel for convolutional operations as there is only a single feature channel in use. Therefore the slightly increased complexity of the Time-CNN makes it slower compared to the MCDCNN in small feature spaces, although this increased complexity led to a more accurate predictor. Of the specialised networks (Temporal CNN, InceptionTime, RNN, Transformer), the Temporal CNN is the fastest architecture by a significant margin.

Given the limited information provided by each individual time series within the dataset, the ability to extract meaningful temporal features would inform the most optimal models. This study demonstrated that the Temporal CNN was the most effective model, performing convolutions along the temporal axis, and extracting valuable feature representations from the sparse spectral information at each timestamp with which to make predictions. Notably, it was able to do this within a much shorter time frame than the InceptionTime, Transformer and RNN networks. Again, the difference in training times between the Temporal CNN, Time-CNN and MCDCNN networks was not small enough to justify the extra performance gained by using a more complex model such as the Temporal CNN. It is again observed that the Transformer model was noticeably slower than the RNN. This is likely down to the number of layers used in the optimal Transformer increasing the number of sequential operations per epoch, and that a Transformer model with fewer Transformer Encoder layers would have a more comparative training time. The lack of parallelisation in the RNN model is noticeable here since the model is comparatively slow against the Temporal CNN whilst containing relatively few parameters, highlighting the strength of parallelisation in convolutional layers.

The RNN was the worst-performing network on this dataset, a stark contrast to the performance seen on the TiSeLaC dataset. This highlights the potential of CNNs for SITS classification in situations where few features are provided but labelled training data is plentiful. In the general case though, the RNN can be expected to scale poorly in high data dimensions due to the costly and complicated training procedure. RNNs, even with contemporary components such as GRUs still have much longer processing times than Temporal CNNs, in general, \cite{SainteFareGarnot2020}. This behaviour is observed even on the relatively short temporal sequences involved within this study, clearly displaying some of the main drawbacks of a pure RNN network for SITS classification. The RNN had per-class accuracies significantly lower than that of the Temporal CNN and InceptionTime networks. It also displayed significant deficiencies in the under-sampled classes as can be seen in Table \ref{tab:SITS-TSI_Per_Class_Accs}. It is interesting to note that in the case of a longer time series, the Transformer model was able to outperform the RNN. As the time series becomes longer, the vanishing gradient deficiency associated with RNNs becomes more pronounced, emphasising the inherent advantages Transformers have on tasks that process longer time series.  

The InceptionTime network is able to utilise a much greater amount of training data due to its complexity, overfitting less than on the TiSeLaC dataset, resulting in a comparably closer performance to the Temporal CNN, being able to accurately classify rarer classes. However, this complexity reduces the scalability of the network, with the training time being prohibitively long when compared to the rest of the surveyed methods. Due to the complexity of the InceptionTime network, the model tends to over-fit slightly on smaller datasets with regularisation techniques having little effect. Decreasing the batch size slightly improves the generalisation of the model but at the cost of drastically increasing the training time. Given the complexity of the InceptionTime network and the muted effect of regularisation techniques, exposing it to more data and variety during training would be the optimal approach for improving performance and subsequently reducing overfitting. This was shown to be the case in the introductory paper for InceptionTime \cite{IsmailFawaz2020} which trained the architecture on the SITS-TSI dataset \cite{Tan2017}, with experiments demonstrating a positive relationship between increasing amounts of training data and accuracy, with linearly increasing training times, 

It was also demonstrated that the relatively simple neural network architectures of the Time-CNN and MCDCNN were able to outperform the more traditional RNN. The Time-CNN, being an improvement of the MCDCNN yet again demonstrated the benefits of incorporating temporal processors that are more carefully developed. The MCDCNN, lacking an advanced mechanism for processing the temporal dimension beyond basic convolutions performed $\sim$5\% worse than the Time-CNN but still outperformed the RNN by $\sim$1\%. The relatively low number of parameters that the Time-CNN and MCDCNN contain also demonstrates their ability to scale to large data domains.

\begin{table}[ht!]
\centering
\begin{tabular}{ |l||c|c|c|c|c|c|c|  }
 \hline
 & \multicolumn{6}{c|}{\textbf{Per class accuracies for each model - SITS-TSI dataset}} \\
 \hline
 Class & Temporal CNN & Time-CNN & MCDCNN & RNN & InceptionTime & Transformer\\
 \hline
 1 & \textbf{98.08} & 97.97 & 97.63 & 97.00 & 97.79 & 97.44\\
 \hline
 2 & 95.92 & \textbf{96.46} & 95.36 & 96.13 & 95.33 & 96.39\\
 \hline
 3 & \textbf{64.69} & 63.51 & 59.20 & 61.20 & 63.36 & 56.52\\
 \hline
 4 & 62.56 & 57.14 & 57.87 & \textbf{67.46} & 49.23 & 65.96\\
 \hline
 5 & \textbf{58.36} & 56.54 & 56.87 & 52.15 & 56.82 & 52.56\\
 \hline
 6 & 80.99 & 79.10 & 80.11 & 78.45 & 77.43 & \textbf{84.37}\\
 \hline
 7 & 80.41 & \textbf{81.79} & 74.51 & 75.56 & 76.22 & 80.37\\
 \hline
 8 & 73.43 & 71.25 & 69.53 & 73.72 & \textbf{78.03} & 73.23\\
 \hline
 9 & 80.62 & 80.77 & 80.76 & 78.49 & \textbf{83.69} & 78.00\\
 \hline
 10 & \textbf{96.03} & 93.37 & 93.81 & 95.74 & 95.89 & 94.98\\
 \hline
 11 & \textbf{73.33} & 60.53 & 69.57 & 70.24 & 70.00 & 62.39\\
 \hline
 12 & 80.57 & \textbf{83.31} & 78.05 & 79.33 & 82.91 & 79.02\\
 \hline
 13 & \textbf{76.54} & 63.03 & 65.75 & 68.39 & 75.14 & 68.97\\
 \hline
 14 & 93.34 & 92.86 & 92.69 & 93.04 & \textbf{93.82} & 93.35\\
 \hline
 15 & 89.73 & \textbf{91.06} & 90.37 & 85.06 & 85.62 & 85.34\\
 \hline
 16 & 50.00 & 41.03 & 28.95 & 45.00 & 45.65 & \textbf{54.17}\\
 \hline
 17 & \textbf{97.47} & 96.29 & 96.30 & 95.82 & 97.24 & 95.31\\
 \hline
 18 & 96.90 & 94.47 & 97.06 & 95.83 & \textbf{97.29} & 96.80\\
 \hline
 19 & 94.34 & \textbf{94.45} & 92.43 & 92.47 & 93.35 & 88.59\\
 \hline
 20 & 84.34 & 78.17 & \textbf{84.96} & 82.14 & 81.54 & 82.68\\
 \hline
 21 & \textbf{80.00} & 61.11 & 40.74 & \textbf{80.00} & 72.73 & 55.56\\
 \hline
 22 & 67.74 & 73.68 & 43.48 & 55.56 & \textbf{100.00} & 51.61\\
 \hline
 23 & 82.22 & \textbf{92.31} & 92.11 & 85.37 & 91.89 & 91.67\\
 \hline
 24 & 75.00 & \textbf{85.71} & 72.73 & 57.14 & 66.67 & 46.15\\
 \hline
\end{tabular}
\caption[SITS-TSI per-class performance results]{\label{tab:SITS-TSI_Per_Class_Accs}Best accuracies on the SITS-TSI dataset for each method, per class.}
\end{table}

In contrast to the results on the TiSeLaC dataset, there is a much narrower range in the strength of performance between the strongest model (Temporal CNN) and weakest model (RNN) on the SITS-TSI dataset. Accordingly, the distribution of the highest accuracy in each class is broader. Each model was the best performer in at least one class, with the Temporal CNN achieving the highest performance in 8 classes, Time-CNN on 7, InceptionTime on 5, and the Transformer and RNN jointly in 2 (the RNN sharing the best performance in class 21 with the Temporal CNN), and the MCDCNN being the best performer only on class 20. The lowest observed accuracy was on class 16 by the MCDCNN, only classifying it correctly 28.95\% of the time, compared to 54.17\% by the Transformer. Given the degree to which some classes are under-sampled and the similarities between the temporal and spectral profiles of the different classes, Despite prevalent class imbalances, every model was largely able to extract features well enough to represent under-sampled classes. Even classes 20-24 that had significantly few training samples commonly display accuracies exceeding 70\%. There was a much higher rate of misclassifications in this dataset than in the TiSeLaC dataset due to class imbalances and this dataset contains more crop-type classes which have similar temporal and spectral attributes. Classes 3 and 5 were commonly misclassified as the other, with class 4 also routinely classified as 3 or 5. 11 was misclassified over several classes, attributed to the few training samples available to create reliable feature maps from. Class 16 which also had few training samples was regularly classified as 13 or 8. Classes 20-24 which are all severely under-sampled were routinely misclassified with 20 misidentified as 19, 21 as either 7,8, or 9, 22 as 3, 4, or 5, 23 as 11, and finally 24 as both 23 and 4. Few significant cases of misclassification occurred for classes with sufficient numbers of training samples and unique temporal and spectral features (e.g. non-crop). These common misclassifications demonstrate the challenging nature of this dataset, with many classes not providing enough samples and features for the various models to extract reliable feature extractors. 

The results on this dataset largely reflect those found on the TiSeLaC dataset, though overall accuracies were lower on this dataset. This is due to the added challenge of a longer temporal axis, reduced spectral features at each timestamp, and a plethora of more classes with similar features and severe class imbalances. Again, the optimised Temporal CNN architecture was able to outperform all other architectures, with the shortest training time of the specialised networks by a considerable margin. Hyperparameter optimisation had a rather muted effect on the performance of the Temporal CNN, with larger dropout rates having the greatest adverse effect on performance. This is in contrast to the experimental results of the TiSeLaC dataset in Appendix \ref{secA2} where hyperparameter changes had modest impacts on performance. Inherently, when there are fewer spectral features per timestamp, performance was largely indifferent to the architectural configuration, and thus a less complex model could be chosen, reducing training times. These results were able to demonstrate that in a drastically different dataset, the Temporal CNN was still able to outperform other the other surveyed methods.

\section{Conclusion and Future Work}
\label{conclusion}

Temporal CNNs are extensively here explored for use in SITS classification on two publicly available datasets. In this paper, an extensive study was conducted to validate the leading performance of Temporal CNNs for SITS classification tasks. This study was broken down into replicating the findings of Pelletier et al. \cite{Pelletier2019} who originally introduced Temporal CNNs for SITS classification, and then surveying the model against a plethora of baseline models that have previously seen success in multi-channel temporal classification tasks.
The set of architectural experiments on the Temporal CNN found that in datasets with dense feature spaces, model tuning has a noted effect on performance, with the dropout rate being the most effective tool to regulate overfitting. Emphasis was placed on how to design architectures that can leverage the spectral and temporal dimensions within the studied datasets. These experiments were carried out on two SITS datasets that consist of vastly different feature spaces, allowing for a simultaneous investigation into the adaptability of Temporal CNNs for extracting features across diverse feature spaces. 

During the testing of the Temporal CNN architecture across the two datasets, it was discovered that Temporal CNNs achieved superior performances over each of the benchmark methods, achieving accuracies of 95.02\% and 87.3\% on the TiSeLaC and SITS-TSI datasets respectively. The  InceptionTime network was the most competitive model compared to the Temporal CNN for both datasets but was the most expensive model to train. These results confirmed the ability of Temporal CNNs to effectively utilise the spectral and temporal dimensions when applying convolutional operations. The impact of the architectural configuration on performance such as the width and depth was studied was both datasets. The influence of regularisation mechanisms and temporal filter size was also examined. Many of these factors were not heavily affected by the feature space of the dataset used except that of filter size. It was found that model performance is partially dependent on the degree to which the temporal filter fits the observed sequential data. It was also shown that unnecessarily high dropout values that appeared to control over-fitting degenerated the overall performance on the test data and that more moderate values are appropriate. The TiSeLaC dataset was more sensitive to architectural optimisations due to having more spectral features at each timestamp compared to the SITS-TSI dataset.

Overall, these results demonstrate that Temporal CNNs provide an effective and efficient method of time series representation on two studied SITS datasets that contained disparate feature spaces, spatial resolution, and temporal length. The experimental architecture demonstrated a remarkable ability to scale up to larger data domains whilst simultaneously exhibiting minimal over-fitting. These results are in agreement with that of previous studies \cite{bai2018empirical,Pelletier2019}, in which Temporal CNNs can be expected to outperform canonical RNNs and simpler CNN architectures for diverse SITS classification tasks. Accordingly, this study suggests that Temporal CNNs should be considered as a more appropriate starting model for deep learning applications for sequential data in place of RNNs \cite{Zhong2019}. The more contemporary models in the form of InceptionTime and the Transformer were also shown to be outperformed by the Temporal CNN in terms of performance and training time. This somewhat agrees with previous studies \cite{russwurm2019breizhcrops, Russwurm2020} where the Temporal CNN has outperformed InceptionTime and was competitive with the Transformer for pre-processed SITS datasets. The Temporal CNN had a more marked advantage against each of the surveyed methods on the less challenging TiSeLaC dataset. Whilst still being the strongest performer on the SITS-TSI dataset, the lead over competing methods was notably reduced. This would imply that with reduced spectral guidance, a longer time series and pronounced class imbalances, the advantages of the Temporal CNN are diminished.

Future work for this paper would entail a deeper analysis of the mechanisms and regularisation procedures explored here. A more challenging dataset comprised of a much longer temporal axis would be useful for exploring how the effectiveness of temporal convolutions fares at scale. There would also be a focus on modifying the network to utilise the spatial dimension of SITS data as has been done recently \cite{Ji2018}. Developing such a model would however restrict the scope of available applicable datasets as spatial information is not routinely provided. By incorporating this feature, models will be able to learn more discriminant features as there is a high correlation in land cover usage for neighbouring pixels. Alternatively, research could be done into making Temporal CNNs more robust to noise within SITS datasets such as cloud cover and salt-and-pepper noise. Transformer models have already been demonstrated to be strong performers on raw multi-spectral datasets \cite{Russwurm2020}, reducing their need for excessive pre-processing steps, and saving significant time and costs. Recent transformer-based models have been shown to outperform Temporal CNNs on SITS classification tasks when classifying at a per-parcel level. Accordingly, it would naturally follow to conduct this comparison on a per-pixel level to confirm if those findings are still consistent.

Applying any recent developments from transfer learning would also help to facilitate accelerated model development for a plethora of SITS classification tasks on a variety of architectures. A multitude of studies has been conducted that release pre-trained models on SITS classification tasks, especially in the domain of crop classification \cite{russwurm2019breizhcrops, Yuan2021}. Transfer learning would have the benefits of reducing the need for large-scale labelled data, extensive pre-processing steps and improved generalisation \cite{Fawaz2019}. A current barrier to the wide-scale usage of transfer learning for SITS tasks is the lack of standardisation in datasets reducing the plausible scope of available pre-trained models due to divergences in dataset compositions. This lack of standardisation in dataset composition and preprocessing steps often means that pre-trained models are incompatible with many available datasets, resulting in severe performance degradation. Overcoming this requires a community effort to create a convention for standardising the composition and pre-processing used when creating public SITS datasets and making them as accessible as possible with accompanying documentation on usage.

\bibliographystyle{unsrt}  
\bibliography{paper}  

\appendix


\section{Supplementary details on deep learning principles}\label{secA1} 

The purpose of this appendix is to provide additional details to principals relating to deep learning.

Deep learning is a subset of machine learning, in which any neural network with three or more layers is considered a DNN \cite{dey2019intelligent}. Several computational layers are concatenated together, with each layer taking inputs from the previous layer. This is known as a \emph{feed-forward} network. These neural networks attempt to simulate the human brain by "learning" to perform a pre-defined task from a large amount of data by learning representations of the data at various levels of abstraction \cite{dey2019intelligent,lecun2015deep}.   

\begin{figure}[ht]
\centering
\includegraphics[width=0.5\textwidth,height=\textheight,keepaspectratio]{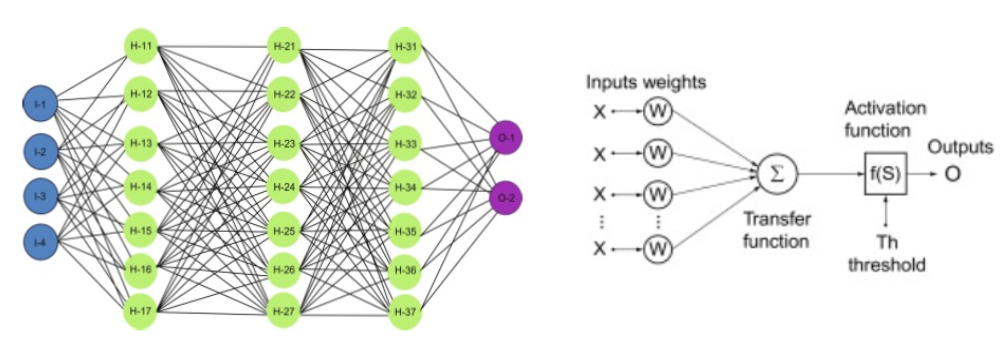}
\caption[DNN architecture and artificial neuron]{(L) feed forward DNN with three hidden layers. (R) a typical artificial neuron. Image sourced from \cite{dey2019intelligent}.}
\label{fig:Example_DNN}
\end{figure}

In Figure~\ref{fig:Example_DNN} , on the left is an example DNN whereby the neurons in blue represent the inputs, the neurons in green represent neurons in the hidden layers and the purple neurons are the outputs. Each layer is comprised of a certain number of units, or namely neurons \cite{lecun2015deep}. The number of input neurons is dependent on the dimension of the instances in the data, whereas the number of neurons in the output layer is comprised of C neurons for a classification task of C classes \cite{Pelletier2019}. In a regression problem, only a single output neuron is needed. The number of hidden layers used and the number of neurons in each are to be selected by the practitioner and is heavily task-dependent.

Looking at the image on the right of Figure~\ref{fig:Example_DNN} is an example of a typical artificial neuron that is at the core of the neural network's learning process. The outputs of a layer $l$ and the activation map $A^{[l]}$ are obtained via a two-step calculation. First, the outputs of the neurons from the previous layer are received as a linear combination. This linear combination of inputs and respective weights is then passed to a non-linear activation function such as tanh, ReLU \cite{krizhevsky2012imagenet}. This can be more formally written as \cite{Pelletier2019}:
\begin{equation}\label{eq:network}
    A^{[l]} = g^{[l]}(W^{[l]}A^{[l-1]}+b^{[1]}), 
\end{equation}
where $W^{[l]}$ and $b^{[l]}$ are the weights and biases of the layer $l$ respectively. Both the weights and biases are learnt by the model during training. 

The activation function as represented by $g^{[l]}$ introduces non-linear combinations of features within the network. If only linear functions are used, then the final output would simply be a linear combination of the inputs. This could be achieved with just a single layer, making the depth of the network largely irrelevant \cite{Pelletier2019}. Correspondingly, the activation function used within this study is the Rectified Linear Units (ReLU) function that has seen popular use within deep learning applications \cite{krizhevsky2012imagenet,Pelletier2019}. It is calculated as: ReLU($z$) = max(0, $z$), which outputs the input value $z$ if it positive, else it outputs 0.

By stacking several specialised hidden layers together in a neural network, the ability of the network to represent complex functions increases. This is enabled by the use the non-linear activations that are used to form a pattern of active hidden units, subsequently allowing the layers to contain a relatively small number of units, reducing the size of the network overall \cite{Pelletier2019,dey2019intelligent}. Section 5.3 explores the experiments conducted surrounding the optimal model width and depth.

The act of training a neural network relates to minimising a given  cost function by finding values of $W = \{W^{[l]}\}_{\forall l}$ and $b = \{b^{[l}\}_{\forall l}$ \cite{Pelletier2019}. The cost function is used to gauge the fit of the model to the data provided. This process is known as empirical risk minimisation and it is done when errors are back-propagated through the network whilst training occurs, updating the values for the weights of each input such that appropriate values will activate the neuron. The set of weights and biases which are used to uniquely identify a particular output is called the feature kernel and they are randomly initialised at model creation \cite{dey2019intelligent}. The cost function $\mathcal{L}$ used is generally defined as the average of the errors made on each individual training instance \cite{Pelletier2019}:
\begin{equation}\label{eq:loss_func}
    \mathcal{J}(\textbf{W}, \textbf{b}) = \frac{1}{n} \sum_{\hat{x_i}} \mathcal{L}(\hat{y_i}, y_i).
\end{equation}
where $\hat{y_i}$ corresponds to the predictions made by the network.

In multi-class classification problems, the loss function $\mathcal{L}(\hat{y_i}, y_i)$ generally used is the categorical crossentropy loss function \cite{zhang2018generalized} and is defined as follows \cite{Pelletier2019}:
\begin{equation}\label{eq:cross_entropy}
    \mathcal{L}(\hat{y_i}, y_i) = -\sum_{y \in \{1,...,C\}} 1 \{y=y_i\}log(p(y \vert x_i)
\end{equation}
\begin{equation}
    = -log(p(y_i \vert x_i))
\end{equation}
where $p(y_i \vert x_i)$ is a representation of the probability that the model predicts the actual class $y_i$ of instance $i$ in the last layer of the network. For networks using categorical cross-entropy as the loss function, the last layer will be a Softmax layer that contains scores for each possible output class and returns the class with the highest score \cite{Pelletier2019}.\\

Now that the core components of a DNN have been introduced, the benefits of neural networks can be fully understood. However, the designing of effective networks requires considerable expertise when deciding factors such as the architecture configuration, related hyperparameter values and optimisation techniques \cite{Pelletier2019}. With appropriate decisions made for each of these, the reward is a model that can learn effective features from the data rather than requiring manual feature engineering \cite{krizhevsky2012imagenet}. Developing models with such large numbers of parameters however creates issues with over-fitting, the phenomena where the model learns the noise within the training data and subsequently fails to generalise to unseen testing data \cite{zhang2021understanding}. Measures to mitigate this issue were experimented with extensively within this study.

\section{Additional results from Temporal CNN experiments}\label{secA2} 

This appendix is concerned with documenting any extra results of interest generated during the research of the optimal Temporal CNN architecture. 

\begin{table*}[ht!]
 \footnotesize
    \centering
    \begin{tabular}{|l|l|l|l|l|l|l|}
    \hline
        \multicolumn{7}{|c|}{Results for finding the optimal width and depth on the SITS-TSI dataset}\\ \hline
        NB\_CONV\_LAYERS & NB\_CONV\_UNITS & NB\_FC\_UNITS & BATCH\_SIZE & DROPOUT & FILTER\_SIZE & OA \\ \hline
        2 & 128 & 256 & 256 & 0.2 & 5 & 87.05 \\ \hline
         &  &  &  &  &  &  \\ \hline
        3 & 64 & 256 & 256 & 0.15 & 5 & 87.08 \\ \hline
        3 & 64 & 256 & 256 & 0.2 & 5 & 86.95 \\ \hline
        3 & 64 & 256 & 256 & 0.5 & 5 & 84.03 \\ \hline
        3 & 128 & 256 & 128 & 0.25 & 5 & 86.84 \\ \hline
        3 & 128 & 256 & 128 & 0.2 & 5 & 87.16 \\ \hline
        3 & 128 & 256 & 128 & 0.25 & 5 & 86.88 \\ \hline
        3 & 128 & 256 & 128 & 0.3 & 5 & 86.88 \\ \hline
        3 & 128 & 512 & 128 & 0.3 & 5 & 87.02 \\ \hline
        3 & 256 & 128 & 128 & 0.2 & 5 & 86.87 \\ \hline
        3 & 256 & 256 & 128 & 0.2 & 5 & 87.13 \\ \hline
        3 & 256 & 256 & 128 & 0.25 & 5 & 86.94 \\ \hline
        3 & 256 & 256 & 128 & 0.3 & 5 & 86.8 \\ \hline
         &  &  &  &  &  &  \\ \hline
        4 & 128 & 256 & 128 & 0.2 & 5 & 87.11 \\ \hline
        4 & 256 & 256 & 128 & 0.35 & 5 & 86.92 \\ \hline
         &  &  &  &  &  &  \\ \hline
        5 & 128 & 256 & 128 & 0.2 & 5 & 87.1 \\ \hline
    \end{tabular}
    \caption[Full Temporal CNN SITS-TSI architecture experimental results]{\label{tab:Full_SITS-TSI_Archi_Experiments}Full list of experiments ran when finding the optimal architecture of the Temporal CNN for the SITS-TSI dataset.}
\end{table*}

\begin{table*}[t!]
\footnotesize
    \centering
    \begin{tabular}{|l|l|l|l|l|l|l|}
    \hline
        \multicolumn{7}{|c|}{Filter size experiments on TiSeLaC dataset}\\ \hline
        NB\textunderscore CONV\textunderscore LAYERS & NB\textunderscore CONV\textunderscore UNITS & NB\textunderscore FC\textunderscore UNITS & BATCH\textunderscore SIZE & DROPOUT & FILTER\textunderscore SIZE & OA \\ \hline
        3 & 128 & 256 & 128 & 0.2 & 3 & 93.94 \\ \hline
        3 & 128 & 256 & 64 & 0.3 & 3 & 94.72 \\ \hline
        3 & 128 & 256 & 128 & 0.182 & 5 & 94.37 \\ \hline
        3 & 128 & 256 & 128 & 0.2 & 5 & 95.02 \\ \hline
        3 & 128 & 256 & 128 & 0.2 & 7 & 93.93 \\ \hline
        3 & 128 & 512 & 128 & 0.182 & 7 & 93.39 \\ \hline
    \end{tabular}
    \caption[Full Temporal CNN TiSeLaC filter size experimental results]{\label{tab:Full_TiSeLaC_Filter_Experiments}Full list of experiments ran when finding the optimal filter size of the Temporal CNN for the TiSeLaC dataset.}
\end{table*}

\begin{table*}[t!]
\footnotesize
    \centering
    \begin{tabular}{|l|l|l|l|l|l|l|}
    \hline
    \multicolumn{7}{|c|}{Results for finding the optimal width and depth on the TiSeLaC dataset}\\ \hline
    NB\textunderscore CONV\textunderscore LAYERS & NB\textunderscore CONV\textunderscore UNITS & NB\textunderscore FC\textunderscore UNITS & BATCH\textunderscore SIZE & DROPOUT & FILTER\textunderscore SIZE & OA \\ \hline
    2 & 64 & 128 & 128 & 0.2 & 5 & 91.64 \\ \hline
    2 & 64 & 256 & 128 & 0.2 & 5 & 92.62 \\ \hline
    2 & 64 & 512 & 128 & 0.2 & 5 & 93.16 \\ \hline
    2 & 128 & 256 & 128 & 0.2 & 5 & 93.28 \\ \hline
    2 & 256 & 256 & 128 & 0.2 & 5 & 94.13 \\ \hline
     &  &  &  &  &  &  \\ \hline
    3 & 64 & 64 & 128 & 0.2 & 5 & 90.9 \\ \hline
    3 & 64 & 128 & 128 & 0.2 & 5 & 92.64 \\ \hline
    3 & 64 & 256 & 128 & 0.2 & 5 & 92.81 \\ \hline
    3 & 64 & 512 & 128 & 0.2 & 5 & 91.33 \\ \hline
    3 & 128 & 128 & 128 & 0.2 & 5 & 93.57 \\ \hline
    3 & 128 & 256 & 128 & 0.182 & 5 & 93.37 \\ \hline
    3 & 128 & 256 & 128 & 0.2 & 5 & 95.02 \\ \hline
    3 & 128 & 256 & 128 & 0.2 & 5 & 94.24 \\ \hline
    3 & 128 & 256 & 64 & 0.182 & 5 & 94.05 \\ \hline
    3 & 128 & 512 & 128 & 0.25 & 5 & 94.02 \\ \hline
    3 & 256 & 256 & 128 & 0.182 & 5 & 94.22 \\ \hline
    3 & 256 & 256 & 128 & 0.2 & 5 & 94.58 \\ \hline
    3 & 512 & 256 & 128 & 0.2 & 5 & 93.48 \\ \hline
    3 & 512 & 512 & 128 & 0.2 & 5 & 93.36 \\ \hline
    3 & 512 & 1024 & 128 & 0.2 & 5 & 93.02 \\ \hline
    3 & 512 & 2048 & 128 & 0.2 & 5 & 93.54 \\ \hline
    3 & 1024 & 256 & 128 & 0.2 & 5 & 93.83 \\ \hline
     &  &  &  &  &  &  \\ \hline
    4 & 64 & 256 & 128 & 0.2 & 5 & 92.8 \\ \hline
    4 & 64 & 512 & 128 & 0.2 & 5 & 92.58 \\ \hline
    4 & 128 & 256 & 128 & 0.2 & 5 & 93.02 \\ \hline
    4 & 128 & 512 & 128 & 0.2 & 5 & 94 \\ \hline
    4 & 256 & 256 & 128 & 0.2 & 5 & 94.44 \\ \hline
    4 & 256 & 256 & 128 & 0.182 & 5 & 93.37 \\ \hline
     &  &  &  &  &  &  \\ \hline
    5 & 64 & 256 & 128 & 0.2 & 5 & 91.62 \\ \hline
    5 & 128 & 256 & 128 & 0.2 & 5 & 91.94 \\ \hline
    5 & 128 & 256 & 128 & 0.3 & 5 & 92.91 \\ \hline
    5 & 256 & 256 & 128 & 0.2 & 5 & 92.94 \\ \hline
    \end{tabular}
    \caption[Full Temporal CNN TiSeLaC architecture experimental results]{\label{tab:Full_TiSeLaC_Archi_Experiments}Full list of experiments ran when finding the optimal architecture of the Temporal CNN for the TiSeLaC dataset.}
\end{table*}

\end{document}